\documentclass[10pt,journal,compsoc]{IEEEtran}
\pdfoutput=1
\usepackage{subfig}
\usepackage{url}
\usepackage{graphicx}
\usepackage{amssymb}
\usepackage{amsthm} 
\usepackage{amsmath}
\usepackage{epsfig}
\usepackage[outdir=./images/]{epstopdf}
\graphicspath{{./images/}}
\usepackage{color}
\usepackage{algorithm}
\usepackage{algorithmic}
\usepackage{array,booktabs}
\usepackage{multirow} 
\usepackage{mathtools}

\newcommand{\real}{\mathbb{R}}

\begin{document}
\title{Elastic Functional Coding of Riemannian Trajectories}

\author{Rushil~Anirudh,~Pavan~Turaga,~Jingyong Su,
       Anuj Srivastava
\IEEEcompsocitemizethanks{\IEEEcompsocthanksitem R. Anirudh and P. Turaga are with the School of Arts, Media, \& Engineering and Department of Electrical, Computer, and Energy Engineering at Arizona State University, Tempe,
AZ. E-mail: \{ranirudh,pturaga\}@asu.edu
\IEEEcompsocthanksitem  Jingyong Su is with the Department of Mathematics \& Statistics at Texas tech University, Lubbock, TX. E-mail: jingyong.su@ttu.edu 
\IEEEcompsocthanksitem  Anuj Srivastava is with the Department of Statistics at Florida State University, Tallahasse, FL. E-mail:anuj@stat.fsu.edu
 \protect\\

}
\thanks{}}
\thispagestyle{plain}

\IEEEtitleabstractindextext{
\begin{abstract}
Visual observations of dynamic phenomena, such as human actions,  are often represented as sequences of smoothly-varying features . In cases where the feature spaces can be structured as Riemannian manifolds, the corresponding representations become trajectories on manifolds. Analysis of these trajectories is challenging due to non-linearity of underlying spaces and high-dimensionality of trajectories. In vision problems, given the nature of physical systems involved, these phenomena are better characterized on a low-dimensional manifold compared to the space of Riemannian trajectories. For instance, if one does not impose  physical constraints of the human body, in data involving human action analysis, the resulting representation space will have highly redundant features. 
Learning an effective, low-dimensional embedding for action representations will have a huge impact in the areas of search and retrieval, visualization, learning, and recognition. Traditional manifold learning addresses this problem for static points in the Euclidean space, but its extension to Riemannian trajectories is non-trivial and remains unexplored. The difficulty lies in inherent non-linearity of the domain and temporal variability of actions that can distort any traditional metric between trajectories. To overcome these issues, we use the framework based on transported square-root velocity fields (TSRVF); this framework has several desirable properties, including a rate-invariant metric and vector space representations. We propose to learn an embedding such that each action trajectory is mapped to a single point in a low-dimensional Euclidean space, and the trajectories that differ only in temporal rates map to the same point. We utilize the TSRVF representation, and accompanying statistical summaries of Riemannian trajectories, to extend existing coding methods such as PCA, KSVD and Label Consistent KSVD to Riemannian trajectories or more generally to Riemannian functions. We show that such coding efficiently captures trajectories in applications such as action recognition, stroke rehabilitation, visual speech recognition, 
clustering and diverse sequence sampling. Using this framework, we obtain state-of-the-art recognition results, while reducing the dimensionality/complexity by a factor of $100 −-250\times$. Since these mappings and codes are invertible,  they can also be used to interactively visualize Riemannian trajectories and synthesize actions.

\end{abstract}
}
\maketitle


\IEEEraisesectionheading{\section{Introduction}\label{sec:introduction}}

\IEEEPARstart{T}here have been significant advances in understanding differential geometric properties of image and video features in vision and robotics. Examples include activity recognition \cite{TuragaCVPR2009,chaudhrycvpr2009,VemulapalliCVPR2014}, medical image analysis \cite{Fletcher2004}, and shape analysis \cite{Srivastava2005}. Some of the popular non-Euclidean features used for activity analysis include shape silhouettes on the Kendall's shape space \cite{ashokPAMI05}, pairwise transformations of skeletal joints on $SE(3)\times SE(3) \dots \times SE(3)$ \cite{VemulapalliCVPR2014}, representing the parameters of a linear dynamical system as points on the Grassmann manifold \cite{TuragaCVPR2009}, and histogram of oriented optical flow (HOOF) on a hyper-sphere \cite{chaudhrycvpr2009}. A commonly occurring theme in many applications is the need to \emph{represent, compare, and manipulate} such representations in a manner that respects certain constraints. 

\begin{figure}[!htb]
\centering
  \includegraphics[clip = true,trim=0mm 0mm 50mm 0mm,width = 3.4in]{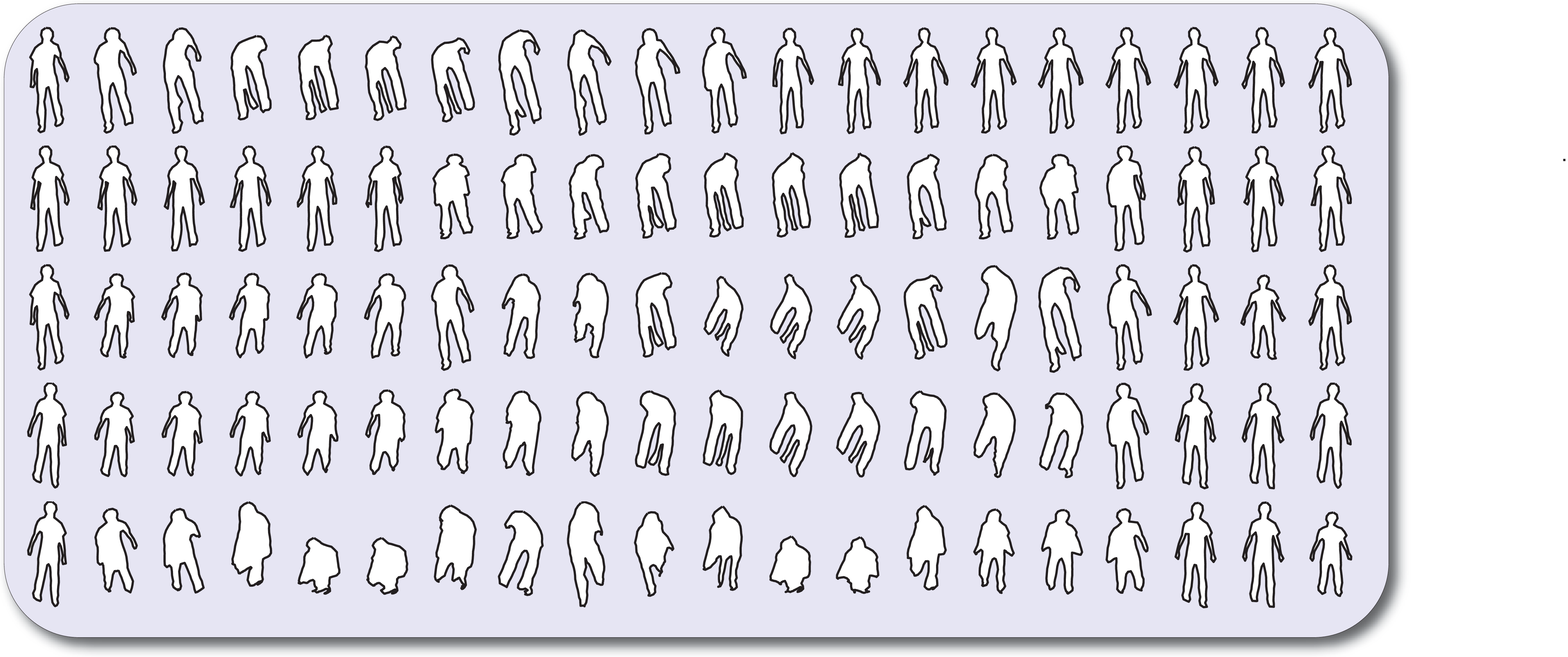}
  \label{fig:mean_warp}
\caption{\footnotesize{Row wise from top -- $S_1$, $S_2$, Warped action $\tilde{S_2}$, Warped mean, Unwarped mean. The TSRVF can enable more accurate estimation of statistical quantities such as average of two actions $S_1, S_2$.}}
\label{fig:test}
\vspace{-0.2in}
\end{figure}

One such constraint is the geometry of such features, since they do not obey conventional Euclidean properties. Another constraint for temporal data such as human actions is the need for speed invariance or \emph{warping}, which causes two sequences to be mis-aligned in time inducing unwanted distortions in the distance metric. Figure \ref{fig:test} shows the effects of ignoring warping, in the context of human actions. Accounting for warping reduces the intra-class distance and improves the inter-class distance. Consequently, statistical quantities such as the \emph{mean sequence} are distorted as seen in figure \ref{fig:test} for two actions $S_1$ and $S_2$. Such effects can cause significant performance losses when using building class templates, without accounting for the changes in speed. The most common way to solve for the mis-alignment problem is to use dynamic time warping (DTW) which originally found its use in speech processing \cite{berndt1994DTW}. For human actions, \cite{Veeraraghavan06,zhouCVPR2012} address this problem using different strategies for features in the Euclidean space. However, DTW behaves as a similarity measure instead of a true distance metric in that it does not naturally allow the estimation of statistical measures such as mean and variance of action trajectories. We seek a representation that is highly discriminative of different classes while factoring out temporal warping to reduce the variability within classes, while also enabling low dimensional coding at the sequence level. 

\begin{figure}[!htb]
\centering
  \includegraphics[clip = true,trim=0mm 0mm 0mm 0mm,height = 2.8in]{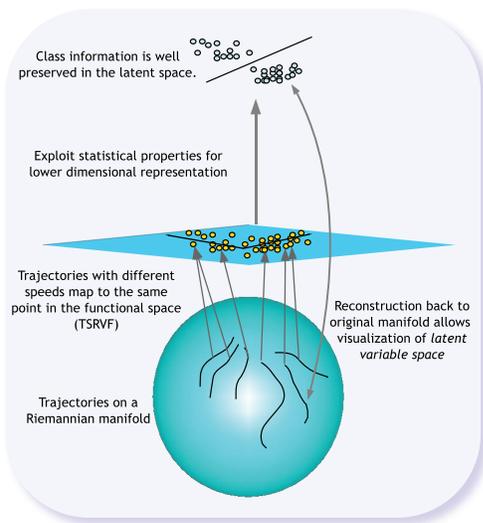}
  \caption{\small{Dimensionality Reduction for Riemannian Trajectories}}
\label{fig:intro_figure}
\vspace{-5pt}
\end{figure}

Learning such a representation is complicated when the features extracted are non-Euclidean (i.e. they do not obey conventional properties of the Euclidean space). Finally, typical representations for action recognition tend to be extremely high dimensional in part because the features are extracted per-frame and stacked. Any computation on such non-Euclidean trajectories can become very easily involved. For example, a recently proposed skeletal representation \cite{VemulapalliCVPR2014} results in a $38220$ dimensional vector for a $15$ joint skeletal system when observed for $35$ frames. Such features do not take into account, the physical constraints of the human body, which translates to giving varying degrees of freedom to different joints. It is therefore a reasonable assumption to make that the \emph{true} space of actions is much lower dimensional. This is similar to the argument that motivated manifold learning for image data, where the number of observed image pixels maybe extremely high dimensional, but the object or scene is often considered to lie on a lower dimensional manifold. A lower dimensional embedding will provide a robust, computationally efficient, and intuitive framework for analysis of actions. In this paper, we address these issues by studying the statistical properties of trajectories on Riemannian manifolds to extract lower dimensional representations or codes. We propose a general framework to {\em code} Riemannian trajectories in a speed invariant fashion that generalizes to many manifolds, the general idea is presented in figure \ref{fig:intro_figure}. We validate our work on three different manifolds - the Grassmann manifold, the product space $SE(3) \times \dots \times SE(3)$, and the space of SPD matrices.

Elastic representations for Riemannian trajectories is relatively new and the lower dimensional embedding of such sequences has remained unexplored. We employ the transport square-root velocity function (TSRVF) representation $-$ a recent development in statistics \cite{Jingyong2014}, to provide a warp invariant representation to the Riemannian trajectories. The TSRVF is also advantageous as it provides a functional representation that is Euclidean. Exploiting this we propose to learn the low dimensional embedding with a Riemannian functional variant of popular coding techniques.  In other words, we are interested in parameterization of Riemannian trajectories, i.e. for $N$ actions $A_i(t), i = 1 \dots N$, our goal is to learn $\mathcal{F}$ such that $\mathcal{F}(x) = A_i$ where $x \in \mathbb{R}^k$ is the set of parameters. Such a model will allow us to compare actions by simply comparing them in their parametric space with respect to $\mathcal{F}$, with significantly faster distance computations, while being able to reconstruct the original actions. In this work, we learn two different kinds of functions using PCA and dictionary learning, which have attractive properties for recognition and visualization.
\vspace{5pt}
\\
\noindent{\bf Broader impact:} While one advantage of embedding Riemannian trajectories into a lower dimensional space is the low cost of storage and transmission, perhaps the biggest advantage is the reduction in complexity of search and retrieval in the latent spaces. Although this work concerns itself primarily with recognition and reconstruction, it is easy to see the opportunities these embeddings present in search applications given that the search space dimension is now $\sim 250 \times$ smaller. We conclusively demonstrate that the embeddings are as discriminative as their original features, therefore guaranteeing an accurate and fast search. The proposed coding scheme also enables visualization of highly abstract properties of human movement in an intuitive manner. We show results on a stroke rehabilitation project which allows us to visualize the \emph{quality} of movement for stroke survivors. These ideas present a lot of opportunity towards building applications that provide users with feedback, while facilitating rehabilitation. We summarize our contributions next.

\vspace{5pt}

\noindent \textbf{Contributions}
\begin{enumerate}
\item An elastic vector-field representation for Riemannian trajectories by modeling the TSRVF on the Grassmann manifold, the product space of $SE(3)\times. .\times SE(3)$ and the space of symmetric positive definite matrices (SPD).
\item Dimensionality reduction for Riemannian trajectories in a speed invariant manner, such that each trajectory is mapped to a single point in the low dimensional space.
\item We present results on three coding techniques that have been generalized for Riemannian Functionals (RF) - PCA, KSVD \cite{AharonEB2006} and Label Consistent KSVD \cite{JiangLD2013}.
\item We show the application of such embedded features or codes in three applications -  action recognition, visual speech recognition, and stroke rehabilitation outperforming all comparable baselines, while being nearly $100 - 250 \times$ more compressed. Their effectiveness is also demonstrated in action clustering and diverse action sampling.
\item The low dimensional codes can be used for visualization of Riemannian trajectories to explore the \emph{latent space} of human movement. We show that these present interesting opportunities for stroke rehabilitation.
\item We perform a thorough analysis of the TSRVF representation testing its stability under different conditions such as noise, length of trajectories and its impact on convergence. 
\end{enumerate}

\subsection{Organization}
A preliminary version of this work appeared in \cite{AnirudhCVPR15}, with application to human activity analysis. In this work we generalize the idea significantly, by considering new applications, new coding methods, and an additional manifold. We also provide a detailed discussion on design choices, parameters and potential weaknesses. We begin in section \ref{sec:related} with a review of related techniques and ideas that provide more context to our contributions. In section \ref{sec:math} we describe the mathematical tools and geometric properties of the various manifolds considered in this work. Section \ref{sec:tsrvf} introduces the TSRVF and discusses its speed invariance properties. In section \ref{sec:rfc}, we propose the algorithm to perform functional coding, specifically for PCA, K-SVD \cite{AharonEB2006}, and Label Consistent K-SVD \cite{JiangLD2013}. We experimentally validate our low dimensional codes in section \ref{sec:expt} on three applications - human activity recognition, visual speech recognition and stroke rehabilitation. We also show applications in visualization, clustering, and diverse sampling for Riemannian trajectories. Section \ref{sec:meta_tsrvf} contains experiments that test the stability and robustness of the TSRVF and the coded representations of Riemannian trajectories under conditions such as noise, and different sampling rates. We also report its convergence rates. We conclude our work and present future directions of research in section \ref{sec:conc}.

\section{Related Work}
\label{sec:related}
\subsection{Elastic metrics for trajectories} The TSRVF is a recent development in statistics \cite{Jingyong2014} that provides a way to represent trajectories on Riemannian manifolds such that the distance between two trajectories is invariant to identical time-warpings. The representation itself lies on a tangent space and is therefore Euclidean, this is discussed further in section \ref{subsec:tsrvf}. The representation was then applied to the problem of visual speech recognition by warping trajectories on the space of SPD matrices \cite{JingyongCVPR14}. A more recent work \cite{ZhangSKLS15} has addressed the arbitrariness of the reference point in the TSRVF representation, by developing a purely intrinsic approach that redefines the TSRVF at the starting point of each trajectory. A version of the representation for Euclidean trajectories - known as the Square-Root Velocity Function (SRVF), was recently applied to skeletal action recognition using joint locations in $\real^3$ with promising results \cite{Devanne2014}. We differentiate our contribution as the first to use the TSRVF representation by representing actions as trajectories in high dimensional non-linear spaces. We use the skeletal feature recently proposed in \cite{VemulapalliCVPR2014}, which models each skeleton as a point on the space of $SE(3) \times \dots \times SE(3)$.
Rate invariance for activities has been addressed before \cite{Veeraraghavan06,zhouCVPR2012}, for example \cite{Veeraraghavan06} models the space of all possible warpings of an action sequence. Such techniques can align sequences correctly, even when features are multi-modal \cite{zhouCVPR2012}. However, most of the techniques are used for recognition which can be achieved with a similarity measure, but we are interested in a representation which serves a more general purpose to 1) provide an effective metric for comparison, recognition, retrieval, etc. and 2) provide a framework for efficient lower dimensional coding which also enables recovery back to the original feature space.

\subsection{Low dimensional data embedding}Principal component analysis has been used extensively in statistics for dimensionality reduction of linear data. It has also been extended to model a wide variety of data types. For high dimensional data in $\mathbb{R}^n$, manifold learning (or non-linear dimensionality reduction) techniques \cite{Tenenbaum00ISOMAP,RoweisLLE00} attempt to identify the underlying low dimensional manifold while preserving specific properties of the original space. Using a robust metric, one could theoretically use such techniques for coding, but the algorithms have impractical memory requirements for very high dimensional data of the order of $\sim 10^4-10^5$, they also do not provide a way of reconstructing the original manifold data. For data already lying on a known manifold, geometry aware mapping of SPD matrices \cite{HarandiSH14} constructs a lower-dimensional SPD manifold, and principal geodesic analysis (PGA) \cite{Fletcher2004} identifies the primary geodesics along which there is maximum variability of data points. We are interested in identifying the variability of sequences instead. Recently, dictionary learning methods for data lying on Riemannian manifolds have been proposed \cite{HoXV13,HarandiSSL13} and could potentially be used to code sequential data but they can be expected to be computationally more intensive. Coding data on Riemannian manifolds is still a new idea with some progress in the past few years, for example recently the Vector of Locally Aggregated Descriptors (VLAD) has also been extended recently to Riemannian manifolds \cite{FarakiCVPR2015}. However, to the best of our knowledge, coding Riemannian trajectories has received little or no attention, but has several attractive advantages. 
\vspace{10pt}
\\
\noindent \textbf{Manifold learning of Trajectories:}
Dimensionality reduction for high dimensional time series is still a relatively new area of research, some recent works have addressed the issue of defining  spatial and temporal neighborhoods. For example, \cite{Lewandowski2014} recently proposed a generalization of Laplacian eigenmaps to incorporate temporal information. Here, the neighborhoods are also a function of time, but the final reduction step still involves mapping a single point in the high dimensional space to a single point in the lower dimensional space. Next, the Gaussian process latent variable model (GPLVM) \cite{lawrence2004gaussian} and its variants, are a set of techniques that perform non-linear dimensionality reduction for data in $\mathbb{R}^N$, while allowing for reconstruction back to the original space. However, its generalization to non-linear Riemannian trajectories is unclear, which is the primary concern of this work. Quantization of Riemannian trajectories has been addressed in \cite{AnirudhT14}, which reduces dimensionality but does not enable visualization. Further, there is loss of information which can cause reduction in recognition performance, whereas we propose to reduce dimensionality by exploiting the latent variable structure of the data. Comparing actions in the latent variable space is similar in concept to learning a linear dynamical system \cite{TuragaCVPR2009} for Euclidean data, where different actions can be compared in the parametric space of the model. 

\subsection{Visualization in Biomedical Applications}
A promising application for the ideas proposed here, is in systems for rehabilitation of patients suffering from impairment of their motor function. Typically visual sensors are used to record and analyze the movement, which drives feedback. An essential aspect of the feedback is the idea of decomposing human motion into its individual components. For example, they can be used to understand abstract ideas such as movement quality \cite{Danziger2012}, gender styles \cite{Troje2002} etc. Troje \cite{Troje2002} proposed to use PCA on individual body joints in $\real^3$, to model different styles of the walking motion.  However, they work with data in the Euclidean space, and explicitly model the temporality of movement using a combination of sinusoids at different frequencies. More recently, a study in neuroscience \cite{Danziger2012} showed that the perceived space of movement in the brain is inherently non-linear and that visualization of different movement attributes can help achieve the \emph{most efficient} movement between two poses. This efficient movement is known to be the geodesic in the \emph{pose space} \cite{Biess2007}. The study was validated on finger tapping, which is a much simpler motion than most human actions. In this work, we generalize these ideas by visualizing entire trajectories of much more complicated systems such as human skeletons and show results on the movement data of stroke-patients obtained from a motion-capture based hospital system \cite{Chen2011Stroke}. 

\section{Mathematical Preliminaries}
\label{sec:math}
In this section we will briefly introduce the properties of manifolds of interest namely -- the product space $SE(3) \times \dots \times SE(3)$, the Grassmann manifold, and the space of symmetric positive definite (SPD) matrices. For an introduction to Riemannian manifolds, we refer the reader to \cite{AbsMahSep2008,Boo03}.

\subsection{Product Space of the Special Euclidean Group}
For action recognition, we represent a stick figure as a combination of relative transformations between joints, as proposed in \cite{VemulapalliCVPR2014}. The resulting feature for each skeleton is interpreted as a point on the product space of $SE(3) \times \dots \times SE(3)$. The skeletal representation explicitly models the 3D geometric relationships between various body parts using rotations and translations in 3D space \cite{VemulapalliCVPR2014}. These transformation matrices lie on the curved space known as the Special Euclidean group $SE(3)$. Therefore the set of all transformations lies on the product space of $SE(3) \times \dots \times SE(3)$.

The special Euclidean group, denoted by $SE(3)$ is a Lie group, containing the set of all $4 \times 4$ matrices of the form 
\begin{equation}
P(R,\overrightarrow{v}) = \begin{bmatrix} R & \overrightarrow{v}\\0 & 1 \end{bmatrix},
\end{equation}
where $R$ denotes the rotation matrix, which is a point on the special orthogonal group $SO(3)$ and $\overrightarrow{v}$ denotes the translation vector, which lies in $\mathbb{R}^3$. The $4 \times 4$ identity matrix $I_4$ is an element of $SE(3)$ and is the identity element of the group. The tangent space of $SE(3)$ at $I_4$ is called its Lie algebra -- denoted here as $\mathfrak{se}(3)$. It can be identified with $4\times 4$ matrices of the form\footnote{We are following the notation to denote the vector space ($\xi \in \real^6$) and the equivalent Lie algebra representation ($\widehat{\xi} \in \mathfrak{se}(3)$) as described in p. 411 of \cite{murray1994mathematical}.}
 \begin{equation}
 \widehat{\xi} = \begin{bmatrix}
 \widehat{\omega} & \overrightarrow{v}\\
 0 & 0
 \end{bmatrix} = \begin{bmatrix}
 0  & -\omega_3 & \omega_2 & v_1\\
 \omega_3 & 0 & -\omega1 & v_2\\
 -\omega_2 & \omega_1 & 0 & v_3\\
 0 & 0 & 0 & 0  \\
 \end{bmatrix},
 \end{equation}
 where $\widehat{\omega}$ is a $3 \times 3$ skew-symmetric matrix and $\overrightarrow{v} \in \real^3$. An equivalent representation is $\xi = [\omega_1,\omega_2,\omega_3,v_1,v_2,v_3]^T \in \real^6$. 
For the exponential and inverse exponential maps, we use the expressions provided on p. 413-414 in \cite{murray1994mathematical}, we reproduce them for completeness here. 

\noindent The exponential map is given by
\begin{equation}
\mbox{exp}~\widehat{\xi} = \begin{bmatrix}
 I & \overrightarrow{v}\\
 0 & 1
 \end{bmatrix}~~ \omega = 0~\mbox{and}~\mbox{exp}~\widehat{\xi} = \begin{bmatrix}
 e^{\widehat{\omega}} & A\overrightarrow{v}\\
 0 & 1
 \end{bmatrix}~~ \omega \neq 0,
\end{equation}
where $e^{\widehat{\omega}}$ is given explicitly by the Rodrigues's formula -- $ = I + \frac{\widehat{\omega}}{\lVert\omega\rVert}\mbox{sin}\lVert\omega\rVert +\frac{\widehat{\omega}^2}{\lVert\omega\rVert^2}(1 - \mbox{cos}\lVert\omega\rVert)$, and  $A = I + \frac{\widehat{\omega}}{\lVert\omega\rVert^2}(1 - \mbox{cos}\lVert\omega\rVert) + \frac{\widehat{\omega}^2}{\lVert\omega\rVert^3}(\lVert\omega\rVert - \mbox{sin}\lVert\omega\rVert)$.

\noindent The inverse exponential map is given by  
\begin{equation}
\widehat{\xi} = \mbox{log}\begin{bmatrix} 
R & \overrightarrow{v}  \\
0 & 1
\end{bmatrix} = \begin{bmatrix}
\widehat{\omega} & A^{-1} \overrightarrow{v} \\
0 & 0
\end{bmatrix},
\end{equation}
where $\widehat{\omega} = \mbox{log}R$, and 
\begin{align*}
A^{-1} = I - \frac{1}{2}\widehat{\omega} + \frac{2~\mbox{sin}\lVert\omega\rVert-\lVert\omega\rVert(1+\mbox{cos}\lVert\omega\rVert)}{2\lVert\omega\rVert^2 \mbox{sin}\lVert\omega\rVert}\widehat{\omega}^2~~~ \omega \neq 0,
\end{align*}
when $\omega = 0$, then $A = I$.

Parallel transport on the product space is the parallel transport of the point on component spaces. Let $T_O(SO(3))$ denote the tangent space at $O \in SO(3)$, then the parallel transport of a $W \in T_O(SO(3))$ from $O$ to $I_{3\times 3}$ is given by $O^TW$. For more details on the properties of the special Euclidean group, we refer the interested reader to \cite{murray1994mathematical}.
\subsection{The space of Symmetric Positive Definite (SPD) matrices} 
We utilize the covariance features for the problem of Visual Speech Recognition (VSR). These features first introduced in \cite{Tuzel2006} have become very popular recently due to their ability to model unstructured data from images such as textures and scenes. A covariance matrix of image features such as pixel locations, intensity and their first and second derivatives is constructed to represent the image. As described in \cite{Pennec2006Tensor}, for a rectangular region $R$, let $\{z_k\}_{k=1\dots n}$ be the $d$-dimensional feature vector of the points inside $R$. The covariance matrix for $R$ is given by $C_R = \frac{1}{n-1}\sum_{k=1}^n (z_k - \mu)(z_k-\mu)^T$. The Riemannian structure of the space of covariance matrices is studied as the space of non-singular, symmetric positive definite matrices \cite{Pennec2006Tensor}. Let $\tilde{\mathcal{P}}(d)$ be the space of $d\times d$ SPD matrices and $\mathcal{P}(d)= \{P | P \in \tilde{\mathcal{P}}(d)\mbox{ and } \mbox{det}(P)=1\}.$ The space $\mathcal{P}(d)$ is a well known symmetric Riemannian manifold, it is the quotient of the special linear group $SL(d) = \{G \in GL(d)|\mbox{det}(G) = 1\}$ by its closed subgroup $SO(d)$ acting on the right and with an $SL(d)$ invariant metric \cite{Jost2011}. Although several metrics have been proposed for this space, few qualify as Riemannian metrics, we use the metrics defined in \cite{SuDKLS12} since the expression for parallel transport is readily available. The Lie algebra of $\mathcal{P}(d)$ is $\mathcal{T}_I(\mathcal{P}(d)) = \{A|A^T =A \mbox{ and }\mbox{trace}(A)=0\}$, where $I$ denotes the $d\times d$ identity matrix and the inner product on $\mathcal{T}_I(\mathcal{P}(d))$ is $\langle A,B\rangle = \mbox{trace}(AB^T)$. The tangent space at $P\in \mathcal{P}(d)$ is $\mathcal{T}_P(\mathcal{P}(d)) = \{PA|A\in \mathcal{T}_I(\mathcal{P}(d))\}$ and $\langle PA,PB\rangle = \mbox{trace}(AB^T)$. 
The exponential map is given as $P\in \mathcal{P}(d)$ and $V \in \mathcal{T}_P(\mathcal{P}(d)), \mbox{exp}_P(V) = \sqrt{P e^{2(P^{-1})V}P}$. The inverse exponential map: For any $P_1, P_2 \in \mathcal{P}(d),\mbox{exp}_{P_1}(P_2) = P_1 log\left(\sqrt{P{_1}^{-1}P_{2}^{2}P{_1}^{-1}}\right).$ Finally, for any $P_1,P_2 \in \mathcal{P}(d)$, the parallel transport of $V\in\mathcal{T}_P(\mathcal{P}(d))$ from $P_1\rightarrow P_2$ is $P_2 T_{12}^T B T_{12}$, where $B = P_1^{-1}V, T_{12} = P_{12}^{-1}P_1^{-1}P_2$ and $P_{12} = \sqrt{P_1^{-1}P_2^2P_1^{-1}}.$

\subsection{Grassmann Manifold as a Shape Space} 
To visualize the action space, we also use shape silhouettes of an actor for different activities. These are interpreted as points on the Grassmann manifold. To obtain a shape as a point, we first obtain a landmark representation of the silhouette by uniformly sampling the shape. Let $L = [(x_1,y_1),(x_2 ,y_2) \dots, (x_m, y_m)]$ be an $m \times 2$ matrix that defines $m$ points on the silhouette whose centroid has been translated to zero. The affine shape space \cite{Goodall1999} is useful to remove small variations in camera locations or the pose of the subject. Affine transforms of a base shape $L_{base}$ can be expressed as $L_{affine}(A) = L_{base}A^T$, and this multiplication by a full rank matrix on the right preserves the column-space of the matrix, $L_{base}$. Thus the 2D subspace of $\mathbb{R}^m$ spanned by the columns of the shape, i.e. $span(L_{base})$ is invariant to affine transforms of the shape. Subspaces such as these can be identified as points on a Grassmann manifold \cite{Begelfor2006,TuragaPAMI2011}.

Denoted by, $\mathcal{G}_{k,m-k}$, the Grassmann manifold is the space whose points are $k$-dimensional hyperplanes (containing the origin) in $\mathbb{R}^m$. An equivalent definition of the Grassmann manifold is as follows: To each $k$-plane, $\nu$ in $\mathcal{G}_{k,m-k}$ corresponds a unique $m\times m$ orthogonal projection matrix, $P$ which is idempotent and of rank $k$. If the columns of a tall $m\times k$ matrix $Y$ spans $\nu$ then $YY^T = P$.
Then the set of all possible projection matrices $\mathbb{P}$, is diffeomorphic to $\mathcal{G}$. The identity element of $\mathbb{P}$ is defined as $Q = diag(I_k,0_{m-k,m-k})$, where $0_{a,b}$ is an $a\times b$ matrix of zeros and $I_k$ is the $k\times k$ identity matrix. The Grassmann manifold $\mathcal{G}$(or $\mathbb{P}$) is a quotient space of the orthogonal group, $O(m)$. Therefore, the geodesic on this manifold can be made explicit by lifting it to a particular geodesic in $O(m)$ \cite{SrivastavaSubspace}. Then the tangent, $X$, to the lifted geodesic curve in $O(m)$ defines the velocity associated with the curve in $\mathbb{P}$. The tangent space of $O(m)$ at identity is $o(m)$, the space of $m\times m$ skew-symmetric matrices, $X$. Moreover in $o(m)$, the Riemannian metric is just the inner product of $\langle X_1,X_2\rangle$ = trace($X_1X_2^T$) which is inherited by $\mathbb{P}$ as well.

The geodesics in $\mathbb{P}$ passing through the point $Q$ (at time t = 0) are of the type $\alpha: (-\epsilon,\epsilon) \mapsto \mathbb{P}, \alpha(t) =$ exp($tX$)$Q$exp($-tX$), where $X$ is a skew-symmetric matrix belonging to the set $M$ where 
\begin{equation}
M =
\left\{\begin{bmatrix}
0 & A\\
-A^T & 0
\end{bmatrix}: A \in \mathbb{R}^{k,n-k} \right\}\subset o(m)
\end{equation}
Therefore the geodesic between $Q$ and any point $P$ is completely specified by an $X \in M$ such that exp($X$)$Q$exp(-$X) = P$. We can construct a geodesic between any two points $P_1,P_2\in \mathbb{P}$ by rotating them to $Q$ and some $P\in\mathbb{P}$. Readers are referred to \cite{SrivastavaSubspace} for more details on the exponential and logarithmic maps of $\mathcal{G}_{k,m-k}$.

\section{Rate Invariant Sequence Comparison}
\label{sec:tsrvf}
In this section we describe the Transport Square Root Velocity Function (TSRVF), recently proposed in \cite{Jingyong2014} as a representation to perform warp invariant comparison between multiple Riemannian trajectories. Using the TSRVF representation for human actions, we propose to learn the latent function space of these Riemannian trajectories in a much lower dimensional space. As we demonstrate in our experiments, such a mapping also provides some robustness to noise which is essential when dealing with noisy sensors.\label{subsec:tsrvf}

Let $\alpha$ denote a smooth trajectory on $\mathcal{M}$ and let $\mathbb{M}$ denote the set of all such trajectories: $\mathbb{M} = \{\alpha :[0,1]\mapsto \mathcal{M}|, \alpha \mbox{ is smooth}\}$. Also define $\Gamma$ to be the set of all orientation preserving diffeomorphisms of [0,1]: $\Gamma = \{\gamma \mapsto [0,1]|\gamma(0) = 0,\gamma(1) = 1, \gamma \mbox{ is a diffeomorphism}\}$. It is important to note that $\gamma$ forms a group under the composition operation. If $\alpha$ is a trajectory on $\mathcal{M}$, then $\alpha \circ\gamma$ is a trajectory that follows the same sequence of points as $\alpha$ but at the evolution rate governed by $\gamma$. The group $\Gamma$ acts on $\mathbb{M}$, $\mathbb{M}\times \Gamma \rightarrow \mathbb{M}$, according to $(\alpha,\gamma) = \alpha \circ \gamma.$ To construct the TSRVF representation, we require a formulation for parallel transporting a vector between two points $p,q \in \mathcal{M}$, denoted by $(v)_{p\rightarrow q}$. For cases where $p$ and $q$ do not fall in the cut loci of each other, the geodesic remains unique, and therefore the parallel transport is well defined. 

The TSRVF \cite{Jingyong2014} for a smooth trajectory $\alpha \in \mathbb{M}$ is the parallel transport of a scaled velocity vector field of $\alpha$ to a reference point $c \in M$ according to:
\begin{equation}
h_\alpha(t) = \left\{ \begin{array}{cc}
\frac{\dot{\alpha}(t)_{\alpha(t)\mapsto c}}{\sqrt{|\dot{\alpha}(t)|}} \in T_c(\mathcal{M}), & |\dot{\alpha}(t)| \neq 0 \\
0 \in T_c(\mathcal{M}) & |\dot{\alpha}(t)| = 0 \end{array}
\right.
\end{equation}
where $|~.~|$ denotes the norm related to the Riemannian metric on $\mathcal{M}$ and $T_c(\mathcal{M})$ denotes the tangent space of $\mathcal{M}$ at $c$. Since $\alpha$ is smooth, so is the vector field $h_\alpha$. Let $\mathcal{H} \subset T_c(\mathcal{M})^{[0,1]}$ be the set of smooth curves in $T_c(\mathcal{M})$ obtained as TSRVFs of trajectories in $\mathcal{M}$, $\mathcal{H} = \{h_\alpha | \alpha \in \mathcal{M}\}$.

\noindent {\bf Distance between TSRVFs: } Since the TSRVFs lie on $T_c(\mathcal{M})$, the distance is measured by the standard $\mathbb{L}^2$ norm given by 
\begin{equation}
d_h(h_{\alpha_1},h_{\alpha_2}) = \left(\int_0^1|h_{\alpha_1}(t)-h_{\alpha_2}(t)|^2 \right)^\frac{1}{2}.
\end{equation}
 If a trajectory $\alpha$ is warped by $\gamma$, to result in $\alpha \circ \gamma$ , the TSRVF of the warped trajectory is given by: 
\begin{equation}
h_{\alpha \circ\gamma}(t) = h_\alpha(\gamma(t))\sqrt{\dot{\gamma(t)}}
\end{equation}
The distance between TSRVFs remains unchanged to warping, i.e.
\begin{equation}
d_h(h_{\alpha_1},h_{\alpha_2}) = d_h(h_{\alpha_1\circ\gamma},h_{\alpha_2\circ\gamma}).
\label{eq:invariance}
\end{equation} 
 The invariance to group action is important as it allows us to compare two trajectories using the optimization problem stated next. 
\vspace{5pt}

\noindent{\bf Metric invariant to temporal variability: } Next, we will use $d_h$ to define a metric between trajectories that is invariant to their time warpings. The basic idea is to partition $\mathbb{M}$ using an equivalence relation using the action of $\Gamma$ and then to inherit $d_h$ on to the quotient space of this equivalence relation. Any two trajectories $\alpha_1, \alpha_2$ are set to be equivalent if there is a warping function $\gamma \in \Gamma$ such that $\alpha_1 = \alpha_2 \circ \gamma$. The distance $d_h$ can be inherited as a metric between the orbits if two conditions are satisfied: (1) the action of $\Gamma$ on $\mathbb{M}$ is by isometries, and (2) the equivalence classes are closed sets. While the first condition has already been verified (see Eqn. \ref{eq:invariance}), the second condition needs more consideration. In fact, since $\Gamma$ is an open set (under the standard norm), its equivalence classes are also consequently open. This issue is resolved in \cite{Jingyong2014} using a larger, {\it closed} set of time-warping functions as follows. Define $\tilde{\Gamma}$ to the set of all non-decreasing, absolutely continuous functions, $\gamma:[0,1]\rightarrow[0,1]$ such that $\gamma(0) = 0$ and $\gamma(1) = 1$. This $\tilde{\Gamma}$ is a semi-group with the composition operation. More importantly, the original warping group $\Gamma$ is a {\it dense} subset of $\tilde{\Gamma}$ and the elements of $\tilde{\Gamma}$ warp the trajectories in the same way as $\Gamma$, except that they allow for singularities \cite{Jingyong2014}.  If we define the equivalence relation using $\tilde{\Gamma}$, instead of $\Gamma$, then orbits are closed and the second condition is satisfied as well. This equivalence relation takes the following form. Any two trajectories $\alpha_1, \alpha_2$ are said to be equivalent, if there exists a $\gamma \in \tilde{\Gamma}$ such that $\alpha_1 = \alpha_2 \circ \gamma$. Since $\Gamma$ is dense in $\tilde{\Gamma}$, and since the mapping $\alpha \mapsto (\alpha(0), h_{\alpha})$ is bijective, we can rewrite this equivalence relation in terms of TSRVF as $\alpha_1\sim \alpha_2$, if \textbf{(a.)} $\alpha_1(0) = \alpha_1(0)$, and \textbf{(b.)} there exists a sequence $\{\gamma_k\}\in \Gamma$ such that $\lim_{k\mapsto\infty}h_{\alpha_1\circ\gamma_k} = h_{\alpha_2}$, this convergence is measured under the $\mathbb{L}^2$ metric. In other words two trajectories are said to be equivalent if they have the same starting point, and the TSRVF of one can be time-warped into the TSRVF of the other using a sequence of warpings. We will use the notation $[\alpha]$ to denote the set of all trajectories that are equivalent to a given $\alpha \in \mathbb{M}$. Now, the distance $d_h$ can be inherited on the quotient space, with the result $d_s$ on $\mathbb{M}/\sim$ (or equivalently $\mathcal{H}/\sim$) given by:

{\footnotesize
\begin{eqnarray}\label{eqn:tsrvf}
&& d_s([\alpha_1],[\alpha_2]) \equiv \inf_{\gamma_1,\gamma_2 \in \tilde{\Gamma}}  d_h((h_{\alpha_1},\gamma_1),(h_{\alpha_2,\gamma_2)}) \nonumber\\ 
&=&\hspace{-15pt} \inf_{\gamma_1,\gamma_2 \in \tilde{\Gamma}} \left(\int_0^1\left|h_{\alpha_1}(\gamma_1(t))\sqrt{\dot{\gamma_1}(t)}-h_{\alpha_2}(\gamma_2(t))\sqrt{\dot{\gamma_2}(t)}\right|^2dt\right)^\frac{1}{2}
\end{eqnarray}
}
The interesting part is that we do not have to solve for the optimizers in $\tilde{\Gamma}$ since $\Gamma$ is dense in $\tilde{\Gamma}$ and, for any $\delta>0$, there exists a $\gamma^*$ such that 
\begin{equation}
|d_h(h_{\alpha_1},h_{\alpha_2}o\gamma^*) - d_s([h_{\alpha_1}],[h_{\alpha_2}])|<\delta.
\end{equation}
This $\gamma^*$ may not be unique but any such $\gamma^*$ is sufficient for our purpose. Further, since $\gamma^*\in \Gamma$, it has an inverse that can be used in further analysis. The minimization over $\Gamma$ is solved for using dynamic programming. Here one samples the interval $[0,1]$ using $T$ discrete points and then restricts to only piecewise linear $\gamma$ that passes through the $T\times T$ grid. Further properties of the metric $d_s$ are provided in \cite{Jingyong2014}. 

\vspace{5pt}

\noindent {\bf Warping human actions: }
In the original formulation of the TSRVF \cite{Jingyong2014}, a set of trajectories were all warped together to produce the mean trajectory. In the context of analyzing skeletal human actions, several design choices are available to warp different actions and maybe chosen to potentially improve performance. For example, warping actions per class may work better for certain kinds of actions that have a very different speed profile, this can be achieved by modifying \eqref{eqn:tsrvf}, to use class information. Next, since the work here is concerned with skeletal representations of humans, different joints have varying degrees of freedom for all actions. Therefore, in the context of skeletal representations, it is reasonable to assume that different joints require different constraints on warping functions. While it may be harder to explicitly impose different constraints to solve for $\gamma$, it can be easily achieved by solving for $\gamma$ per joint trajectory instead of the entire skeleton.

\section{Riemannian Functional Coding}
\label{sec:rfc}
A state of the art feature for skeletal action recognition -- the Lie Algebra Relative Pairs (LARP) features \cite{VemulapalliCVPR2014} uses the relative configurations of every joint to every other joints, which provides a very robust representation, but also ends up being extremely high dimensional. For example, for a $15$ joint skeletal system, the LARP representation lies in a $182\times 6$ dimensional space, therefore an action sequence with $35$ frames has a final representation that has $38220$ dimensions. Such features do not encode the physical constraints on the human body while performing different movements because explicitly encoding such constraints may require hand tuning specific configurations for different applications, which may not always be obvious, and is labor intensive. Therefore, for a given set of human actions, if one can identify a lower dimensional \emph{latent variable} space, which automatically encodes the physical constraints, while removing the redundancy in the original feature representation - one can theoretically represent entire actions as lower dimensional points. This is an extension to existing manifold learning techniques to Riemannian trajectories. It is useful to distinguish the lower dimensional manifold of sequences that is being learned from the Riemannian manifold that represents the individual features such as LARP on $SE(3) \times.. \times SE(3)$ etc. Our goal is to exploit the redundancy in these high dimensional features to learn a lower dimensional embedding without significant information loss. Further, the TSRVF representation, provides us speed invariance which is essential for human actions, this results in an embedding where trajectories that only differ in their rates of execution will map to the same point or to points that are very close in the lower dimensional space. 

We study two main applications of coding - 1) visualization of high dimensional Riemannian trajectories, and 2) classification. For visualization, one key property is to be able to reconstruct back from the low dimensional space, which is easily done using principal component analysis (PCA). For classification, we show results on discriminative coding methods such as K-SVD, LC-KSVD, in additional to PCA, that learn a dictionary where each atom is a trajectory. More generally, common manifold learning techniques such as Isomap \cite{Tenenbaum00ISOMAP}, and LLE \cite{RoweisLLE00} can also be used to perform coding, while keeping in mind that it is not easy to obtain the original feature from the low dimensional code. Further, the trajectories tend to be extremely high dimensional (of the order of $10^4 - 10^5$), therefore most manifold learning techniques require massive memory requirements.

Next we describe the algorithm to obtain low dimensional codes using PCA and dictionary learning algorithms.

\begin{algorithm}[!htb]
\caption{Riemannian Functional Coding}
\label{algo:seq_pca}
\begin{algorithmic}[1]
\STATE {\bf Input}:  \small{$\alpha_1(t),\alpha_2(t)\dots \alpha_N(t) \in \mathbb{M}$ }
\STATE {\bf Output}: \small{Codes $C \in \real^{d\times N}$, in a basis $B \in \real^{D\times d}, d<<D$}
\STATE \small{Compute the Riemannian center of mass $\mu(t)$, which also aligns $\tilde{\alpha}_1(t),
\tilde{\alpha}_2(t)\dots \tilde{\alpha}_N(t)$ \cite{Jingyong2014}.}
\FOR{$i\leftarrow [1 \dots N]$}
\FOR{$t\leftarrow[1\dots T]$}
\STATE Compute shooting vectors $v(i,t) \in T_{\mu(t)}(M)$ as $v(i,t) = \mbox{\textbf{exp}}^{-1}_{\mu(t)}(\tilde{\alpha}_i(t))$
\ENDFOR
\STATE Define $V(i) = [v(i,1)^T~v(i,2)^T~ \dots~ v(i,T)^T]^T$
\ENDFOR
\STATE $[C,B] = \mathcal{F}(V)$. // \footnotesize{$\mathcal{F}$ can be any Euclidean coding scheme}
\end{algorithmic}
\end{algorithm}
\vspace{-10pt}
\subsection{Representing an elastic trajectory as a vector field} 
\begin{figure}[!htb]
\centering
  \includegraphics[clip = true,trim=30mm 50mm 40mm 70mm,height = 2in]{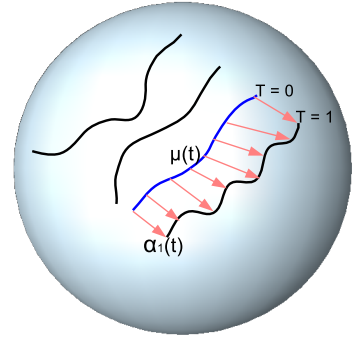}
  \vspace{-15pt}
  \caption{\footnotesize{Representing the warped trajectories on a manifold as a vector field, allows us to use existing algorithms to perform dimensionality reduction efficiently, while also respecting the geometric and temporal constraints.}}
\label{fig:mfpca_algo}
\vspace{-5pt}
\end{figure}

The TSRVF representation allows the evaluation of first and second order statistics on {\em entire sequences of actions} and define quantities such as the variability of actions, which we can use to estimate the redundancy in the data similar to the Euclidean space. We utilize the TSRVF to obtain the ideal warping between sequences, such that the warped sequence is equivalent to its TSRVF. To obtain a low dimensional embedding, first we represent the sequences as deviations from a reference sequence using tangent vectors.  For manifolds such as $SE(3)$ the natural ``origin'' $I_4$ can be used, in other cases the sequence mean \cite{Jingyong2014} by definition lies equi-distant from all the points and therefore is a suitable candidate. In all our experiments, we found the tangent vectors obtained from the mean sequence to be much more robust and discriminative. Next, we obtain the \emph{shooting vectors}, which are the tangent vectors one would travel along, starting from the average sequence $\mu(t)$ at $\tau = 0$ to reach the $i^{th}$ action $\tilde{\alpha}_i(t)$ at time $\tau = 1$, this is depicted in figure \ref{fig:mfpca_algo}. Note here that $\tau$ is the time in the sequence space which is different from $t$, which is time in the original manifold space. The combined shooting vectors can be interpreted as a {\it sequence tangent} that takes us from one point to another in sequence space, in unit time. Since we are representing each trajectory as a vector field, we can use existing algorithms to perform coding treating the sequences as points, because we have accounted for the temporal information. The algorithm \ref{algo:seq_pca} describes the process to perform coding using a generic coding function represented as $\mathcal{F}: \real^D \rightarrow \real^d,$ where $d << D $. In the algorithm, $C$ represents the low dimensional representation in the basis/dictionary $B$ that is learned using $\mathcal{F}$.

\noindent {\bf Complexity:} Computing the mean trajectory and simultaneously warping $N$ trajectories for a single iteration can be done in $\mathcal{O}(N(T^2+\nu))$, where the cost to compute the TSRVF is $\mathcal{O}(\nu)$. If we assume the cost of computing the exponential map is $\mathcal{O}(m)$, algorithm \ref{algo:seq_pca} has a time complexity of $\mathcal{O}(mNT)$. This can be a computational bottle neck for manifolds that do not have a closed form solution for the exponential and logarithmic maps. However, the warping needs to be done once offline, as test trajectories can be warped to the computed mean sequence in $\mathcal{O}(T^2 + \nu)$. Further, both the mean and shooting vector computation can be parallelized to improve speed.
 
\noindent {\bf Reconstructing trajectories from codes:}
If the $\mathcal{F}$ is chosen such that it can be easily inverted, i.e. we can find an appropriate $\mathcal{F}^{-1}:\real^d \rightarrow \real^D$, then the lower dimensional embedding can be used to reconstruct a trajectory on the manifold, $\mathcal{M}$, by traveling along the reconstructed tangents from the mean, $\mu(t)$. This is described in algorithm \ref{algo:pca_recon}.

\begin{algorithm}[!htb]
\caption{Reconstructing Non Euclidean Trajectories}
\label{algo:pca_recon}
\begin{algorithmic}[1]
\STATE {\bf Input}: $C \in \mathbb{R}^{d\times N}, d<<D$, $B \in \mathbb{R}^{D\times d}$, $\mu(t)$.
\STATE {\bf Output}: $\widehat{\alpha}(t) \in \mathbb{M}$
\FOR{$i\leftarrow [1 \dots N]$}
\vspace{5pt}
\STATE $\widehat{V}_i = \mathcal{F}^{-1}(B,C)$
\STATE Rearrange $\widehat{V}_i$ as an $m\times T$ matrix, where $T$ is the length of each sequence.
\FOR{$t\leftarrow[1\dots T]$}
\STATE $\widehat{\alpha}_i(t) = \mbox{\textbf{exp}}_{\mu(t)}(\widehat{V}_i(t),1)$
\ENDFOR
\ENDFOR
\end{algorithmic}
\end{algorithm}
\vspace{-10pt}
\subsection{Choices of coding techniques}
Since the final representation before dimensionality reduction lies in a vector space, any Euclidean coding scheme can be chosen depending on the application. We focus on two main techniques to demonstrate the ideas. First we perform principal component analysis (PCA) since it can be computed efficiently for extremely high dimensional data, it allows reconstruction by which we can obtain the original features, and it also provides an intuitive interpretation to visualize the high dimensional data in $2D$ or $3D$. This version of Riemannian Functional PCA (RF-PCA, previously referred to as mfPCA in \cite{AnirudhCVPR15}), generalizes functional PCA to Riemannian manifolds, and also generalizes principal geodesic analysis (PGA)\cite{Fletcher2004} to sequential data. Next, we use dictionary learning algorithms, allowing us to exploit sparsity. K-SVD \cite{AharonEB2006} is one of the most popular dictionary learning algorithms that has been influential in a wide variety of problems. Recently, label consistent - KSVD (LCKSVD) \cite{JiangLD2013} improved the results for recognition. K-Hyperline clustering \cite{He2009KHypl} is a special case of K-SVD where the sparsity is forced to be $1$, i.e. each point is approximated by a single dictionary atom. It is expected that since K-SVD relaxes the need for the bases to be orthogonal, it achieves much more efficient codes, that are much more compact, have the additional desirable property of sparsity and perform nearly as well as the original features themselves.
\vspace{10pt}

\begin{figure}[!htb]
\centering
  \includegraphics[clip = true,trim=10mm 0mm 10mm 10mm,height = 2.3in]{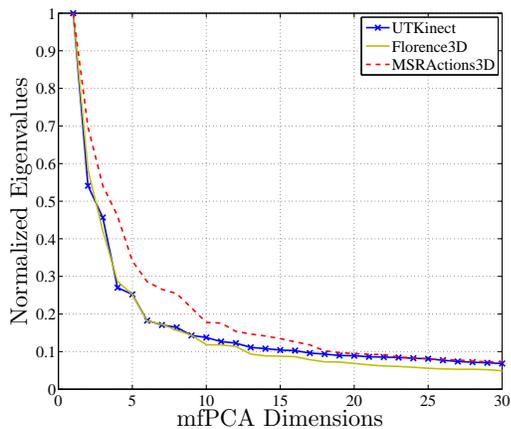}
  \vspace{-5pt}
  \caption{\footnotesize{Eigenvalue decay for MSRActions3D \cite{li2010action}, UTKinect \cite{xia2012view}, and Florence3D \cite{SeidenariVBBP13} datasets obtained with RF-PCA. UTKinect and Florence3D have 10 and 9 different classes respectively, as a result the corresponding eigenvalue decay saturates at around $10$ dimensions. MSRActions consists of $20$ classes and the decay saturates later at around $20$.}}
\label{fig:mfpca_eigen}
\vspace{-20pt}
\end{figure}

\noindent {\bf Eigenvalue decay using RF-PCA:} To first corroborate our hypothesis that Riemannian trajectories are often far lower dimensional than the feature spaces in which they are observed, we show the eigenvalue decay in figure \ref{fig:mfpca_eigen}, after performing RF-PCA on three commonly used datasets in skeletal action recognition. It is evident that most of the variation in the datasets is captured by 10-20 eigenvectors of the covariance matrix. It is also interesting to note that that RF-PCA does a good job of approximating the different classes in the product space of $SE(3) \times \dots \times SE(3)$. The MSRActions dataset \cite{li2010action} contains $20$ classes and correspondingly the eigenvalue decay flattens around $20$. In comparison the UTKinect \cite{xia2012view} and Florence3D \cite{SeidenariVBBP13} datasets contain $10$ and $9$ classes of actions respectively, which is reflected in the eigenvalue decay that flattens closer to around $10$. Features in the RF-PCA tend to be lower dimensional and more robust to noise, which is helpful in reducing the amount of pre/post processing required for optimal performance. 


\section{Experimental Evaluation}
\label{sec:expt}
We evaluate our low dimensional Riemannian coding approach in several applications and show their advantages over conventional techniques that take geometry into account as well as other Euclidean approaches. First we address the problem of activity recognition from depth sensors such as the Microsoft Kinect. We show that a low dimensional embedding can perform as well or better than the high dimensional features on benchmark datasets. Next we evaluate our framework on the problem of visual speech recognition (VSR), or also known as lip-reading from videos. We show that, all other factors remaining the same, our low dimensional codes outperform many baselines. Finally, we also address the problem of movement quality assessment in the context of stroke rehabilitation with state-of-the-art results. We also show that low dimensional mapping provides an intuitive visual interpretation to understand quality of movement in stroke rehabilitation.

\subsection{Action recognition}
We use a recently proposed feature called Lie algebra relative pairs (LARP) \cite{VemulapalliCVPR2014} for skeleton action recognition. This feature maps each skeleton to a point on the product space of $SE(3) \times SE(3) \dots \times SE(3)$, where it is modeled using transformations between joint pairs. It was shown to be very effective on three benchmark datasets - UTKinect \cite{xia2012view}, Florence3D \cite{SeidenariVBBP13}, and MSR Actions3D \cite{li2010action}. We show that using geometry aware warping results in significant improvement in recognition. Further, we show that it is possible to do so with a representational feature dimension that is $250 \times$ smaller than state-of-the-art. 

{\bf Florence3D actions dataset} \cite{SeidenariVBBP13} contains $9$ actions performed by $10$ different subjects  repeated two or three times by each actor. There are 15 joints on the skeleton data collected using the Kinect sensor. There are a total of $182$ relative joint interactions which are encoded in the features. 

{\bf UTKinect actions dataset} \cite{xia2012view} contains 10 actions performed by 10 subjects, each action is repeated twice by the actor. Totally, there are 199 action sequences. Information regarding 20 different joints is provided. There are a total of $342$ relative joint interactions.

{\bf MSRActions3D dataset} \cite{li2010action} contains a total of 557 labeled action sequences consisting of 20 actions performed by 10 subjects. There are 20 joint locations provided on the skeletal data, which gives $342$ relative joint interactions.
\subsubsection{Alternative Representations} 
We compare the performance of our representation with various other recently proposed related methods:

{\bf Lie Algebra Relative Pairs} (LARP): Recently proposed in \cite{VemulapalliCVPR2014}, this feature is shown to model skeletons effectively. We will compare our results to those obtained using  the LARP feature with warping obtained from DTW and unwarped sequences as baselines.

{\bf Body Parts + SquareRoot Velocity Function (BP + SRVF)} : A skeleton is a collection of body parts where each skeletal sequence is represented as a combination of multiple body part sequences, proposed in \cite{Devanne2014}. It is also relevant to our work because the authors use the SRVF for ideal warping, which is the vector space version of the representation used in this paper. The representational dimension is calculated assuming the number of body parts $N_{jp} = 10$, per skeleton\cite{Devanne2014}.

{\bf Principal Geodesic Analysis} (PGA)\cite{Fletcher2004}: Performs PCA on the tangent space of static points on a manifold. We code individual points using this technique and concatenate the final feature vector before classification.
\begin{table}[!htb]
\centering
\begin{tabular}{ |l|p{0.8in}|p{0.6in}| }\hline
Feature &\footnotesize{Representational Dimension} & Accuracy\\\hline
BP+SRVF \cite{Devanne2014}& 30000 &87.04 \\[0.5ex]
LARP \cite{VemulapalliCVPR2014} &38220 & 86.27 \\[0.5ex]
DTW  \cite{VemulapalliCVPR2014} & 38220 &86.74 \\[0.5ex]
PGA \cite{Fletcher2004} & 6370 &79.01 \\[0.5ex]	\hline
TSRVF & 38200 &89.50 \\[0.5ex]
RF-KSVD & {\bf 45} (sparse) &{\bf 88.55} \\[0.5ex]
RF- LCKSVD & {\bf 60} (sparse) &{\bf 89.02} \\[0.5ex]
RF-PCA & {\bf 110} &{\bf 89.67} \\[0.5ex]\hline
\end{tabular}
\vspace{-5pt}
 \caption{\footnotesize{Recognition performance on the Florence3D actions dataset \cite{SeidenariVBBP13} for different feature spaces.}}
 \label{tab:Recog_Florence3D}
\end{table}
\begin{table}[!htb]
\centering
\begin{tabular}{ |l|p{0.8in}|p{0.6in}| }\hline
Feature &\footnotesize{Representational Dimension} & Accuracy\\\hline
BP+SRVF \cite{Devanne2014}& 60000&91.10 \\[0.5ex]
HOJ3D \cite{xia2012view} & N/A&90.92  \\[0.5ex]
LARP \cite{VemulapalliCVPR2014} &151,848 & 93.57 \\[0.5ex]
DTW  \cite{VemulapalliCVPR2014} & 151,848 &92.17 \\[0.5ex]
PGA \cite{Fletcher2004} & 25308 &91.26  \\[0.5ex]	\hline
TSRVF & 151,848 &94.47 \\[0.5ex]
RF-KSVD & {\bf 50} (sparse) &{ 92.67} \\[0.5ex]
RF-LCKSVD & {\bf 50} (sparse) &{\bf 94.87} \\[0.5ex]
RF-PCA & {\bf 105} &{\bf 94.87} \\[0.5ex]\hline

\end{tabular}
 \caption{\small{Recognition performance on the UTKinect actions dataset \cite{xia2012view}.}}
 \label{tab:Recog_UTKinect}
\end{table}
\begin{table}[!htb]
\centering  
\begin{tabular}{ |l|p{0.8in}|p{0.8in}| } \hline
Feature & \footnotesize{Representational Dimension} & Accuracy\\\hline
BP + SRVF \cite{Devanne2014} &60000  & $\mathbf{87.28 \pm 2.99}$ \\[0.5ex]
HON4D \cite{oreifej2013hon4d} &N/A &$82.15 \pm 4.18$\\ [0.5ex]
LARP\cite{VemulapalliCVPR2014} & 155,952&$75.57 \pm 3.43$ \\[0.5ex]
DTW\cite{VemulapalliCVPR2014} &155,952 &$78.75 \pm 3.08$ \\[0.5ex]
PGA \cite{Fletcher2004}& 25,992 &$72.06 \pm 3.12$\\[0.5ex] \hline
TSRVF &155,952 &$84.62 \pm 3.08$ \\[0.5ex]
RF-KSVD &{\bf 120} (sparse) &$84.45 \pm 3.15$ \\[0.5ex]
RF-LCKSVD  &{\bf 50} (sparse) &$83.60 \pm 3.14$ \\[0.5ex]
RF-PCA &{\bf 250} &$\mathbf{85.16 \pm 3.13}$ \\[0.5ex]
\hline
\end{tabular}
\vspace{-5pt}
 \caption{\footnotesize{Recognition performance on the MSRActions3D dataset \cite{li2010action} following the protocol of \cite{oreifej2013hon4d} by testing on 20 classes, with all possible combinations of test train subjects.}}
 \label{tab:Recog_MSRActions}
\end{table}
\subsubsection{Evaluation Settings}
The skeletal joint data obtained from low cost sensors are often noisy, as a result of which post-processing methods such as Fourier Temporal Pyramid (FTP) \cite{WangCVPR2012} have been shown to be very effective for recognition in the presence of noise. FTP is also a powerful tool to work around alignment issues, as it transforms a time series into the Fourier domain and discards the high frequency components. By the nature of FTP, the final feature is invariant to any form of warping. One of the contributions of this work is to demonstrate the effectiveness of geometry aware warping over conventional methods, and then explore the space of these warped sequences, which is not easily possible with FTP. Therefore, we perform our recognition experiments on the non-Euclidean features sequences without FTP. We computed the mean on $SE(3)$ extrinsically for the sake of computation, since the Riemannian center of mass for the manifold is iterative. In general this can lead to errors since the log map for $SE(3)$ is not unique, however we found this to work well enough to model skeletal movement in our experiments. This can easily be replaced with the more stable intrinsic version, for details on implementations we refer the reader to \cite{Duan2013Riemannian}. For classification, we use a one-vs-all SVM classifier following the protocol of \cite{VemulapalliCVPR2014}, and set the $C$ parameter to 1 in all our experiments. For the Florence3D and UTKinect datasets we use five different combinations of test-train scenarios and average the results. For the MSRActions dataset, we follow the train-test protocol of \cite{oreifej2013hon4d} by performing recognition on all 242 scenarios of 10 subjects of which half are used for training, and the rest for testing.

\subsubsection{Recognition results}
The recognition rates for Florence 3D, UTKinect, and MSRActions3D are shown in tables \ref{tab:Recog_Florence3D}, \ref{tab:Recog_UTKinect} and \ref{tab:Recog_MSRActions} respectively. It is clear from the results that using TSRVF on a Riemannian feature, leads to significant improvement in performance. Further, using RF-PCA improves the results slightly, perhaps due to robustness to noise, but more importantly, reduces the representational dimension of each action by a factor of nearly $250$. Sparse codes obtained by K-SVD, and LC-KSVD further reduce the data requirement on the features, where LC-KSVD performs as well as RF-PCA while also inducing sparsity in the codes. The improvements are significant compared to using DTW as a baseline; the performance is around $3\%$ better on Florence3D, $2\%$ on UTKinect, and $7\%$ averaged over all test train variations on MSR Actions 3D. Although BP+SRVF \cite{Devanne2014} has higher recognition numbers on the MSRActions3D, our contribution lies in the significant advantage obtained using the LARP features with RF-PCA (over $7\%$ on average). We observed that simple features in $\mathbb{R}^N$ performed exceedingly well on MSRActions3D, for example using relative joint positions (given by $\overrightarrow{v} = J_1 - J_2$, where $J_1$ and $J_2$ are 3D coordinates joints 1 and 2.) on the MSRActions3D with SRVF and PCA we obtain $87.17 \pm 3.08 \%$ by embedding every action into $\mathbb{R}^{250}\times$, which is similar to \cite{Devanne2014}, but in a much lower dimensional space. The performance of LCKSVD on MSRActions3D is lower than state-of-the-art because it requires a large number of samples per action class to learn a robust dictionary. There are $\sim 20$ action classes in the dataset, but only $557$ actions, therefore we are restricted to learn a much smaller dictionary. In other datasets with enough samples per class, LCKSVD performs as well as RF-PCA while also generating sparse codes.

We also show that performing PCA on the shooting vectors is significantly better than performing PCA on individual time samples using Principal Geodesic Analysis. The dimensions for LARP features are calculated as $6 \times J \times T$, where $J$ is the number of relative joint pairs per skeleton, and $T$ is the number of frames per video. We learn the RF-PCA basis using the training data for each dataset, and project the test data onto the orthogonal basis. 

\subsection{Visual speech recognition}
Next we evaluate our method Visual Speech Recognition (VSR) on the OuluVS database \cite{ZhaoBP09} and show that the proposed coding framework outperforms comparable techniques at a significantly reduced dimensionality. VSR is the problem of understanding  speech as observed from videos. The dataset contains audio and video clues, but we will use only the videos to perform recognition, this problem is also known as automatic lipreading. Speech is a dynamic process, and very much like human movement. It is also subject to significant variation in speed, as a result of which accounting for speed becomes important before choosing a metric between two samples of speech \cite{JingyongCVPR14}. 


\noindent {\bf OuluVS database} \cite{ZhaoBP09}: This includes 20 speakers uttering $10$ phrases: {\it Hello, Excuse me, I am sorry, Thank you, Good bye, See you, Nice to meet you, You are welcome, How are you, Have a good time.} Each phrase is repeated $5$ times. All the videos in the database are segmented, with the mouth regions determined by the manually labeled eye positions in each frame. We compare our results to those reported in \cite{JingyongCVPR14}, who used covariance descriptors on the space of SPD matrices to model the visual speech using TSRVF. There are two protocols of evaluation for VSR typically, speaker independent test and speaker dependent test (SDT). We report results on the latter following \cite{JingyongCVPR14}.
\subsubsection{Feature descriptor and evaluation settings}
We use the covariance descriptor \cite{Tuzel2006} which has proven to be very effective in modeling unstructured data such as textures, materials etc. We follow the feature extraction process as described in \cite{JingyongCVPR14}, to show the effectiveness of our framework. For the covariance descriptor, seven features are extracted including $\{x,y, I(x,y),|\frac{\partial I}{\partial x}|,|\frac{\partial I}{\partial x}|,|\frac{\partial^2 I}{\partial x}|,|\frac{\partial^2 I}{\partial x}| \}$, where $x,y$ are the pixel locations, $I(x,y)$ is the intensity of the pixel, and  the remaining terms are the first \& second partial derivatives of the image with respect to $x,y$. This is extracted at each pixel, within a bounded region around the mouth. These covariance matrices are summed up to obtain a single $7 \times 7$ region covariance descriptor per frame. These form a trajectory of such matrices per video, which we use to calculate its TSRVF and subsequently the low dimensional codes.

We show improved results are achieved while also providing highly compressed feature representations as shown in Table \ref{tab:VSR_expt}. We train a one-vs-all SVM similar to the previous experiment, on the shooting vectors directly, by training on $60\%$ of the subjects for each spoken phrase, this is repeated for all train/test combinations. We obtain an accuracy of $74.05\%$ on uncompressed shooting vectors, as compared to $66.0\%$ using a 1-NN classifier on all the 1000 videos proposed in \cite{JingyongCVPR14}. The functional codes using different coding schemes outperform even the SVM results by around $1.5\%$. While the improvement is not significant, it is important to note that there is a reduction in the feature representation by a factor of nearly $100\times$.

\begin{table}[!htb]
\centering  
\begin{tabular}{ |p{1.0in}|p{0.8in}|c| } \hline
Feature & \footnotesize{Representational Dimension} & Accuracy\\\hline
Cov SPD \cite{Tuzel2006} &2450  & $31.9$ \\[0.5ex]
TSRVF + NN \cite{JingyongCVPR14} &2450  & $66.0$ \\[0.5ex]
Spatio-temporal\cite{ZhaoBP09} &N/A &$70.2$ \footnotesize{(800 videos)}\\ [0.5ex]
PGA \cite{Fletcher2004} &1000  & $72.42 \pm 3.14$ \\[0.5ex]
\hline
TSRVF + SVM &2450  & $74.05 \pm 4.14$ \\[0.5ex]
RF - LCKSVD & {\bf 20} (sparse)  & $74.04 \pm 3.5$ \\[0.5ex]
RF - KSVD &{\bf 20} (sparse)  & $\mathbf{75.63 \pm 4.45}$ \\[0.5ex]
RF - PCA  &30  & $\mathbf{75.3\pm 5.41} $ \\[0.5ex]
\hline
\end{tabular}
\vspace{-5pt}
 \caption{\footnotesize{Visual speech recognition performance on the OuluVS database \cite{ZhaoBP09} on $1000$ videos using the subject dependent testing (SDT). Results show that the functional coding representation outperforms previous state-of-the-art with similar features, while significantly reducing dimensionality.}}
 \label{tab:VSR_expt}
\end{table}

\subsection{Movement quality for stroke rehabilitation}

\begin{figure}[!htb]
\centering
  \includegraphics[height = 2in]{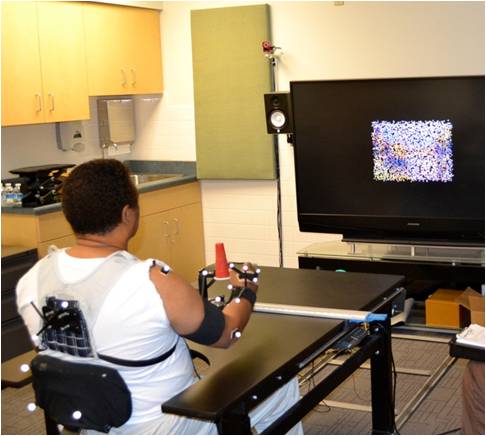}
  \vspace{-5pt}
  \caption{\footnotesize{The stroke rehabilitation system \cite{Chen2011Stroke}, that uses a $14$ marker configuration to provide feedback on motor function for stroke patients. A typical evaluation protocol requires a therapist to observe a specified movement to give a score indicating the quality of movement.}}
\label{fig:stroke}
\vspace{-10pt}
\end{figure}

Each year stroke leaves millions of patients disabled with reduced motor function, which severely restricts a person's ability to perform activities of daily living. Fortunately, the recent decade has seen the development of rehabilitation systems with varying degrees of automated guidance, to help the patients regain a part of their motor function. A typical system is shown in figure \ref{fig:stroke}, which was developed by Chen et al. \cite{Chen2011Stroke}. The system uses $14$ markers to analyze and study the patient's movement (eg. reach and grasp), usually in the presence of a therapist who then provides a \emph{movement quality score}, such as the Wolf Motor Function Test (WMFT) \cite{Wolf2001wmft}.

Our goal in this experiment is to predict the quality of the stroke survivor's movement as well as the therapist, so that such systems can be home-based with fewer therapist interventions. There are 14 markers on the right hand, arm and torso in a hospital setting. A total of $19$ impaired subjects perform multiple repetitions of reach and grasp movements, both on-table and elevated (with the additional force of gravity acting against their movement). Each subject performs $4$  sets of reach and grasp movements to different target locations, with each set having $10$ repetitions. 

\subsubsection{Feature description and evaluation settings}
We choose 4 joints -- back, shoulder, elbow, and wrist. This is used to represent them in relative configurations to each other as is done in LARP \cite{VemulapalliCVPR2014} resulting in each \emph{hand skeleton} that lies in $SE(3)\times \dots \times SE(3)$ as earlier. The problem now reduces to performing logistic regression on trajectories that lie in $SE(3)\times \dots \times SE(3)$. The stroke survivors were also evaluated by the WMFT \cite{Wolf2001wmft} on the day of recording, where a therapist evaluates the subject's ability on a scale of 1 - 5 (with 5 being least impaired to 1 being most impaired). We use these scores as the ground truth, and predict the quality scores using the LARP features extracted from the hand markers. The dataset is small in size due to the difficulty in obtaining data from stroke survivors, therefore we use the evaluation protocol of \cite{VenkataramanCVPRW2013}, where we train on all but one test sample for regression. We compare our results to Shape of Phase Space (SoPS) \cite{VenkataramanCVPRW2013}, who perform a reconstruction of the phase space from individual motion trajectories in each dimension of each joint. 

  
\begin{figure}[!htb]
\centering
  \subfloat[\footnotesize{Easily visualizing quality of movement in RFPCA space}]{ 
  \includegraphics*[clip = true,trim=70mm 40mm 60mm 30mm,height = 1.5in]{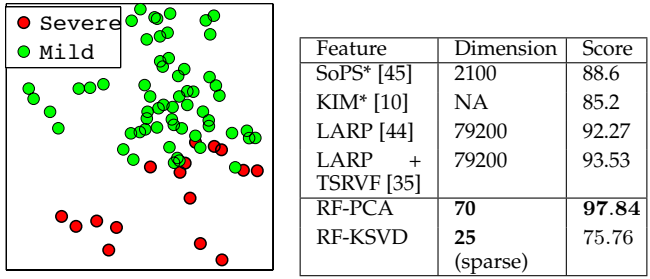}
  \label{fig:mfpca_stroke}}
  \centering
  \footnotesize{\subfloat[\footnotesize{Predicting the quality of movement in the rehabilitation of stroke survivors.}]{
\begin{tabular}[b]{|p{0.55in}|p{0.5in}|p{0.25in}|} \hline
Feature & \footnotesize{Dimension} & \footnotesize{Score}\\\hline
SoPS* \cite{VenkataramanCVPRW2013} &2100 & 88.6 \\[0.5ex]
KIM* \cite{Chen2011Stroke} &NA & 85.2 \\[0.5ex]
LARP \cite{VemulapalliCVPR2014} & 79200& 92.27  \\[0.5ex]
\footnotesize{LARP + TSRVF \cite{Jingyong2014}} & 79200& 93.53 \\[0.5ex]
\hline
RF-PCA &{\bf 70}  & $\mathbf{97.84}$ \\[0.5ex]
RF-KSVD &{\bf 25} (sparse)  & $75.76$ \\[0.5ex]
\hline
\end{tabular}
\label{tab:stroke_expt}}}
\caption{\footnotesize{The RF-PCA is able to accurately predict movement quality as compared to an expert therapist which can improve home-based systems for stroke rehabilitation.}}
\vspace{-5pt}
\end{figure}

 \noindent Table \ref{tab:stroke_expt} shows the results for different features. The baseline, using the features as it is, gives a correlation score of $92.27$ to the therapist's WMFT evaluation.  Adding elasticity to the curves in the $SE(3)$ product space improves the correlation score to $93.53$. The functional codes improves the score significantly to $97.84$, while using only 70 dimensions giving state of the art performance. We also compare our score to the kinematic based features proposed by \cite{VenkataramanCVPRW2013}. 
\noindent \textbf{Visualizing quality:} Next, figure \ref{fig:mfpca_stroke} shows the different movements in the lower dimensional space. Visualizing the movements in RF-PCA space, it is evident that even in $\real^2$, information about the quality of movement is captured. Movements which are indicative of high impairment in the motor function appear to be physically separated from the movements which indicate mild or less impairment. It is easy to see the opportunities such visualizations present for rehabilitation, for example a recent study in neuroscience \cite{Danziger2012} showed that real-time visual feedback can help \emph{learn} the most efficient control strategies for certain movements.

\subsection{Reconstruction and visualization of actions}
We also show results on visualization and exploration of human actions as Riemannian trajectories. Since shapes are easy to visualize, we use the silhouette feature as a point on the Grassmann manifold.

{\bf UMD actions dataset} \cite{Veeraraghavan06}: This is a relatively constrained dataset, which has a static background allowing us to easily extract shape silhouettes. It contains 100 sequences consisting of $10$ different actions repeated $10$ times by the same actor. For this dataset, we use the shape silhouette of the actor as our feature, because of its easy visualization as compared to other non-linear features. 


\begin{figure*}[!htb]
\vspace{-5pt}
\centering
\subfloat[\footnotesize{Row-wise: functional-Precis}]{
  \includegraphics[clip = true,trim=0mm 5mm 0mm 0mm,height = 2.1in]{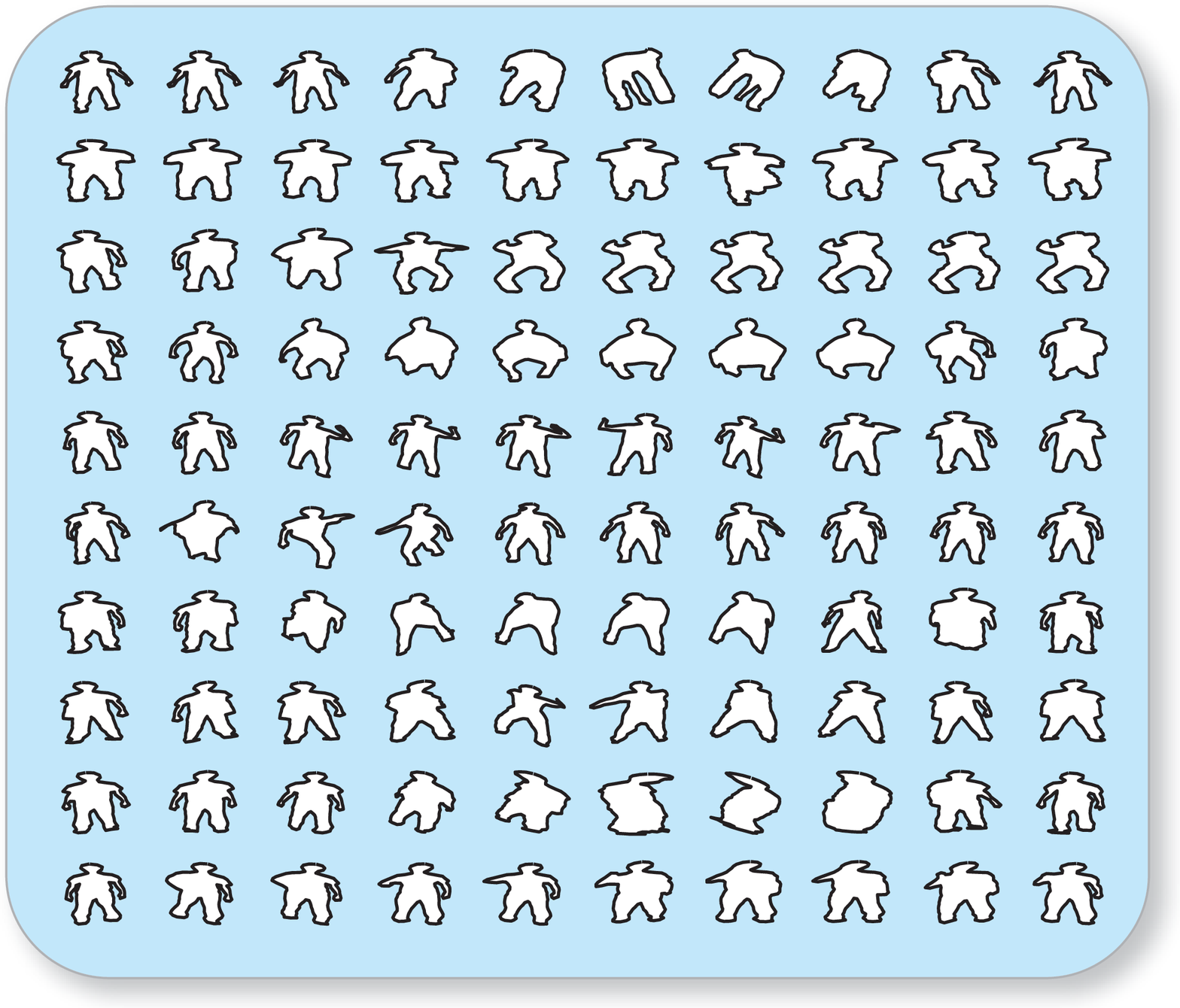}
  \label{fig:PrecisExemplars}}
\subfloat[\footnotesize{Row-wise: functional K-medoids}]{
  \includegraphics[clip = true,trim=0mm 5mm 0mm 0mm,height = 2.1in]{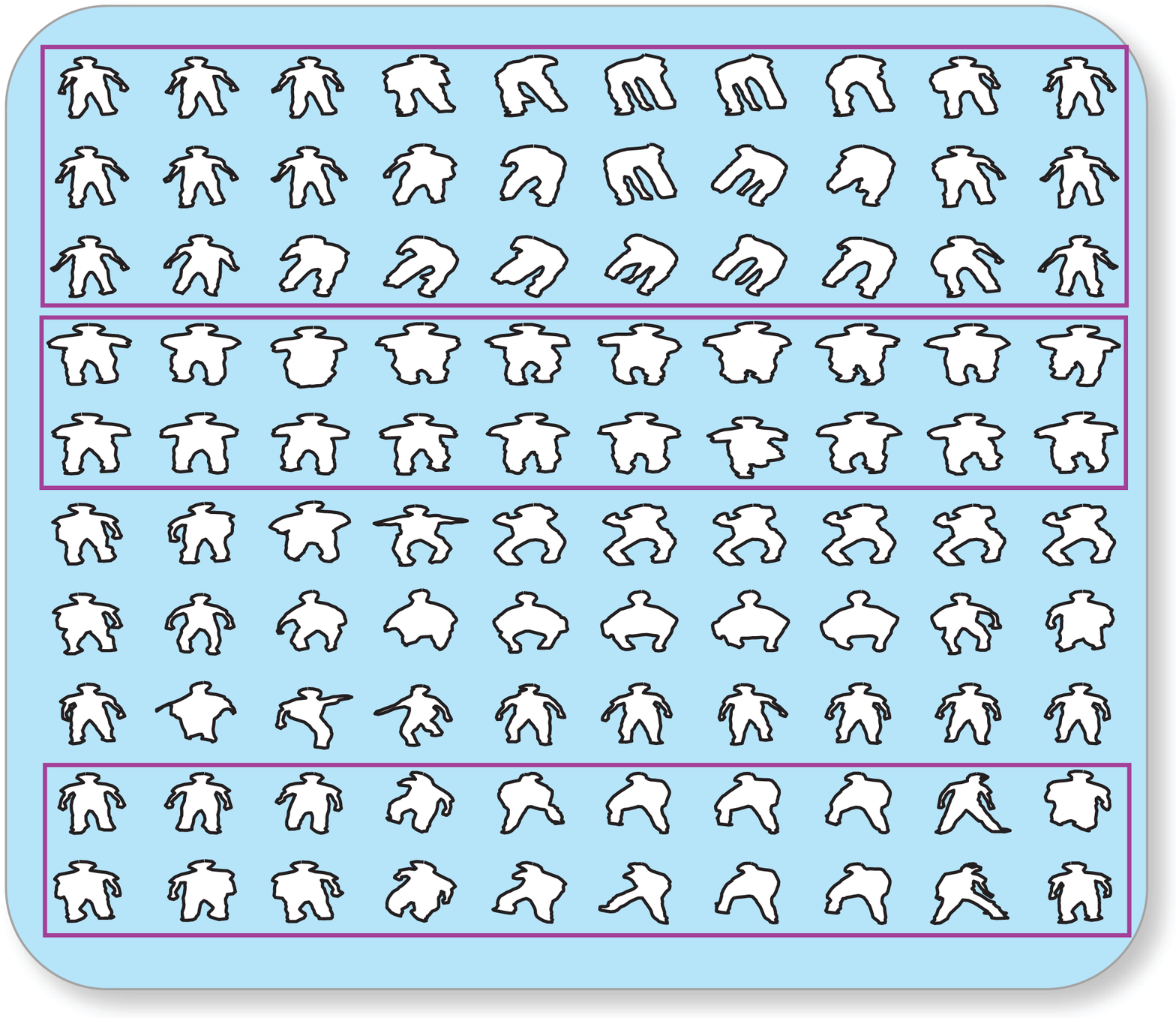}
  \label{fig:KmedoidsExemplars}}
\subfloat[\footnotesize{Exploring the 2D latent action space}]{
  \includegraphics[clip = true,trim=0mm 5mm 0mm 0mm,height = 2.2in,]{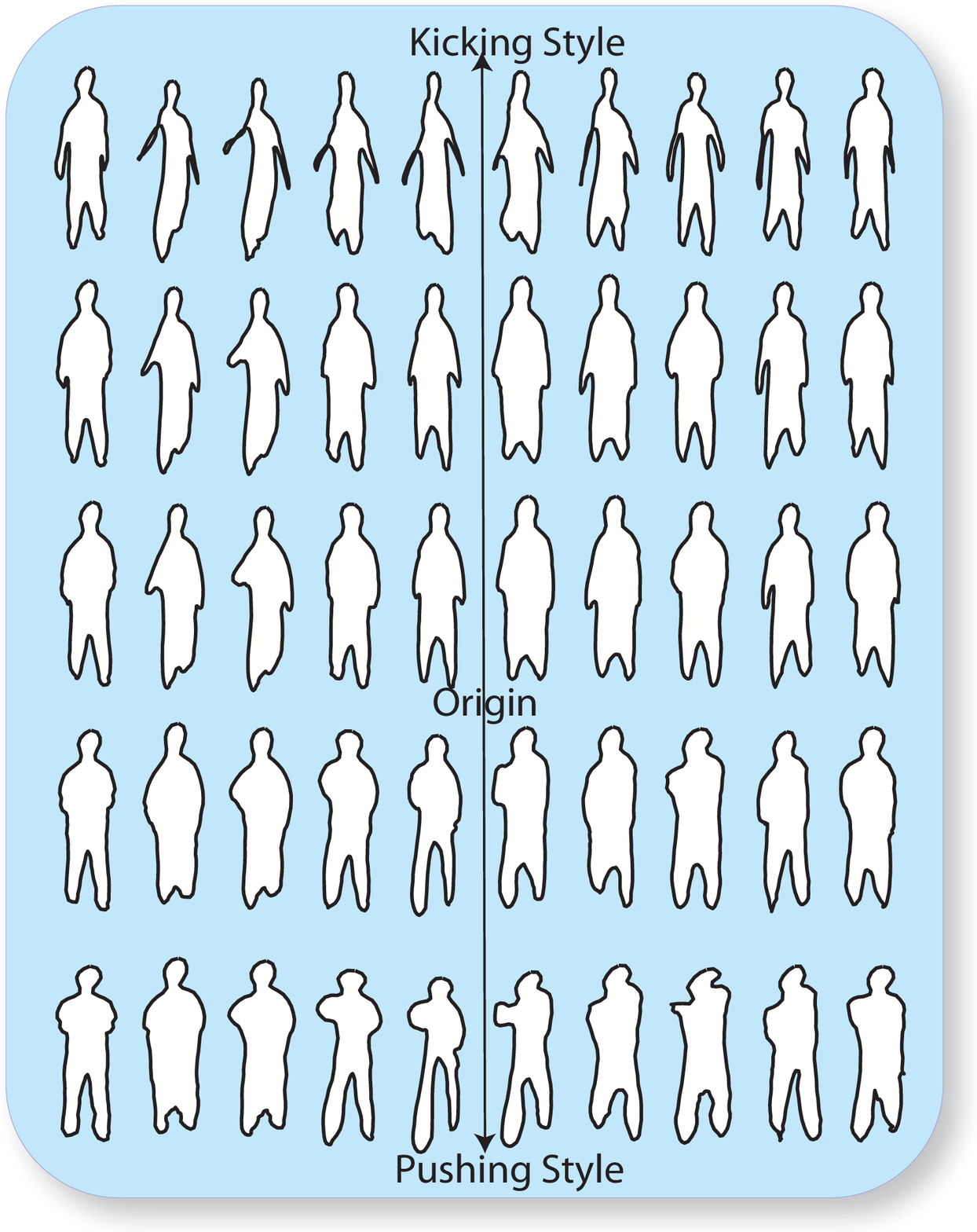}
\label{fig:pca_actions}}
\caption{\footnotesize{Diverse action sampling using Precis\cite{ShroffTC11} by sampling in RF-PCA space $\in \mathbb{R}^{10}$ on a highly skewed dataset. K-medoids picks more samples (marked) from classes that have a higher representation, while Precis remains invariant to it. The K-medoids and diverse clustering operations are performed $\sim 500 \times$ faster in the RF-PCA space. Figure \ref{fig:pca_actions} shows a 2D axis sampled in the latent space. It's clearly seen that even in only 2 dimensions, some action information ("style") is discernible.}}
\label{fig:Seq_Precis}
\vspace{-10pt}
\end{figure*}
\subsubsection{Reconstruction results}
Once we have mapped the actions onto their lower dimensional space using RF-PCA, we can reconstruct them back easily using algorithm \ref{algo:pca_recon}.
We show that high dimensional action sequences that lie in non-Euclidean spaces can be effectively embedded into a lower dimensional latent variable space.
Figure \ref{fig:pca_actions} shows the sampling of one axis at different points. As expected, the ``origin'' of the dataset contains no information about any action, but moving in the positive or negative direction of the axis results in different {\it styles} as shown. Note, that since we are only using $2$ dimensions, there is a loss of information, but the variations are still visually discernible.
\subsection{Diverse sequence sampling}
Next, we show that applications such as clustering can also benefit from a robust distance metric that the TSRVF provides. Further, performing clustering is significantly faster in the lower dimensional vector space, such as the one obtained with RF-PCA. We perform these experiments on the UMD Actions data with actions as trajectories on the Grassmann manifold. K-means for data on manifolds involves generalizing the notion of distance to the geodesic distance and the mean to the Riemannian center of mass. We can further generalize this to sequences on manifolds by replacing the geodesic distance with the TSRVF distance and the mean by the RCM of sequences as defined in \cite{Jingyong2014}. A variant of this problem is to identify the different kinds of groups within a population, i.e. clustering {\it diversly}, which is a harder problem in general and cannot be optimally solved using K-means. Instead we use manifold Precis which is a {\it diverse sampling} method \cite{ShroffTC11}. Precis is an unsupervised exemplar selection algorithm for points on a Riemannian manifold, i.e. it picks a set of $K$ most representative points $S$ from a data set $X$. The algorithm works by jointly optimizing between approximation error and diversity of the exemplars, i.e. forcing the exemplars to be as different as possible while covering all the points. 

To demonstrate the generalizability of our functional codes, we perform an experiment to perform K-means clustering and diverse clustering of \emph{entire sequences}. In the experiment on the UMD actions dataset, we constructed a collection of actions that were chosen such that different classes had significantly different populations in the collection. Action centers obtained with K-medoids is shown in figure \ref{fig:KmedoidsExemplars} and as expected classes which have a higher population are over represented in the chosen samples as compared to Precis (figure \ref{fig:PrecisExemplars}) which is invariant to the distribution. Due to the low dimensional Euclidean representation, these techniques can be easily extended to suit sequential data in a speed invariant fashion due to the TSRVF and at speeds $\sim 500 \times$ faster due to RF-PCA.
\section{Analysis of the TSRVF Representation}
\label{sec:meta_tsrvf}
In this section, we consider different factors that influence the stability and robustness of the TSRVF representation, thereby affecting its coding performance. Factors such as (a) it's stability for different choices of the reference point, (b) the effect of noise on functional coding, and (c) arbitrary length of a trajectory, are realistic scenarios that occur in many applications.

\subsection{Stability to the choice of reference point}
\label{sec:distort}
\begin{figure*}[!ht]
\centering
\subfloat[\footnotesize{Convergence \& choice of reference point {\bf C}}]{
  \includegraphics*[clip = true,trim=20mm 5mm 20mm 10mm,width=0.35\linewidth]{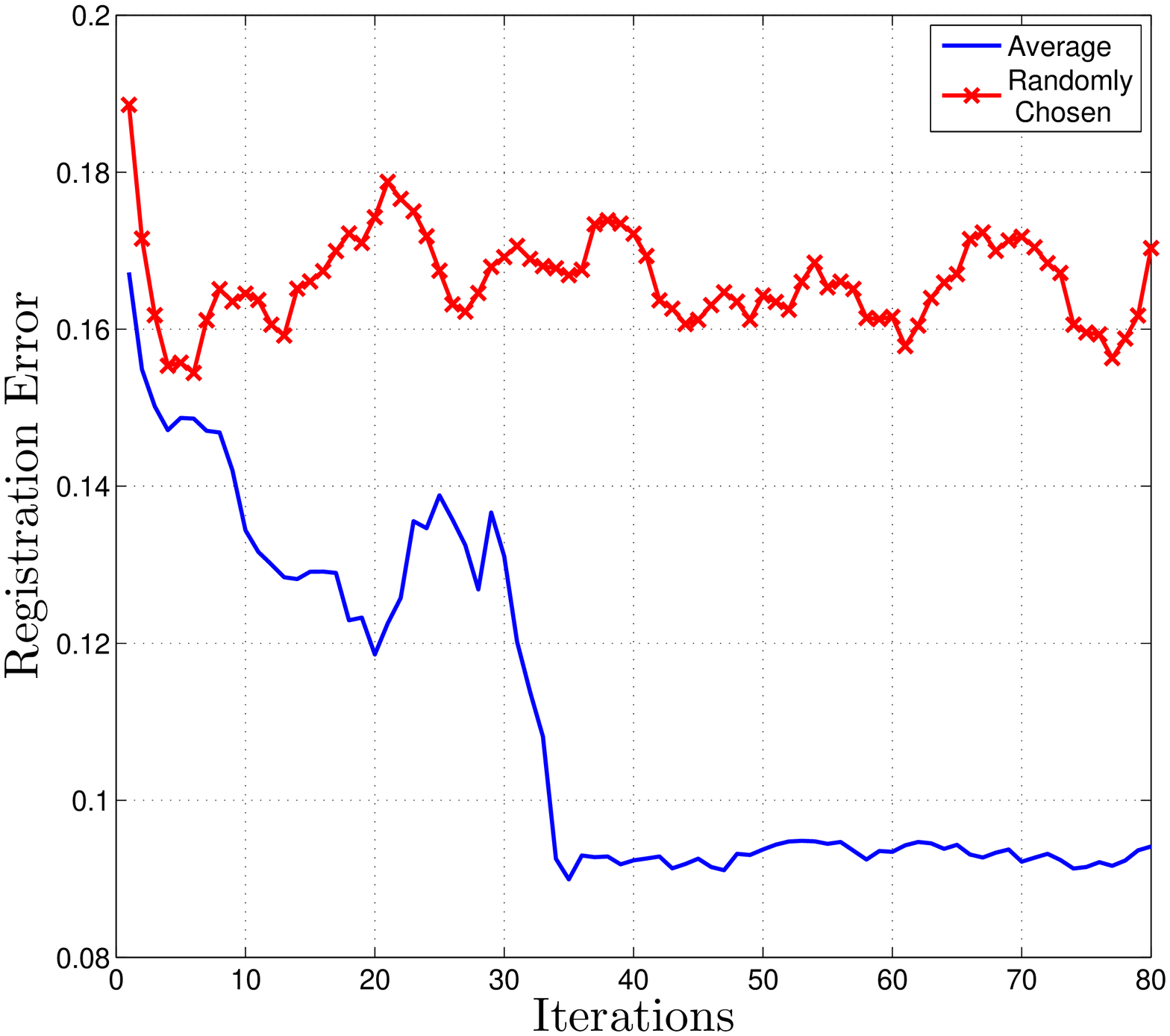}\label{fig:stability}}
\subfloat[\footnotesize{Presence of Noise}]{
  \includegraphics*[clip = true,trim=30mm 0mm 10mm 10mm,width=0.32\linewidth]{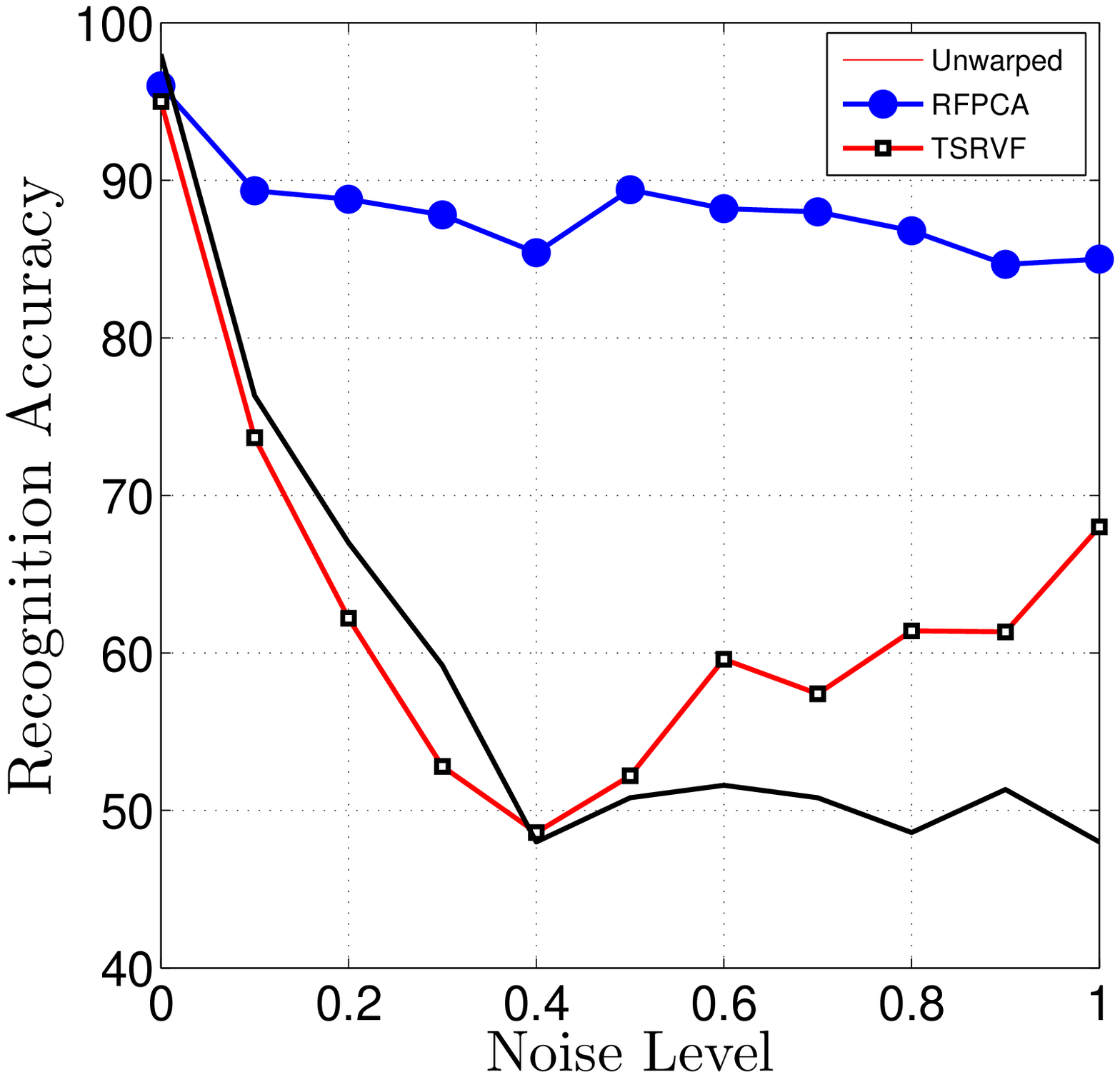}\label{fig:noise}}
\subfloat[\footnotesize{Trajectory length and sampling rate}]{
  \includegraphics*[clip = true,trim=43mm 0mm 15mm 10mm,width=0.3\linewidth]{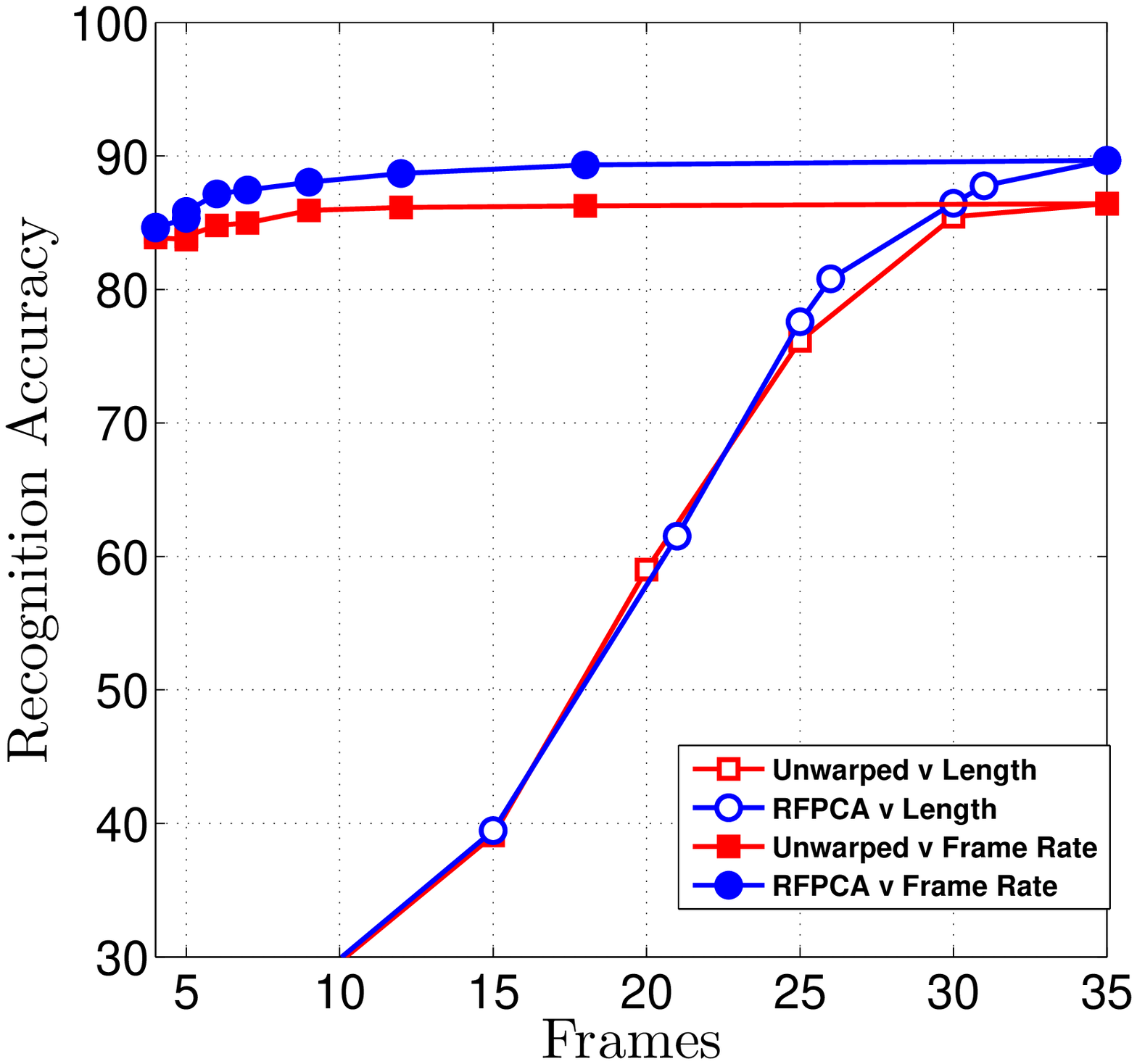}\label{fig:sampling}}
  
\caption{\footnotesize{Robustness experiments for different factors as measured by their effect on recognition accuracy. Experiments in table (\ref{fig:stability}) and figure (\ref{fig:noise}) are performed on the Grassmann manifold, \& figure (\ref{fig:sampling}) shows results on the $SE(3) \times SE(3) \dots SE(3)$ manifold. It can be clearly seen that the RFPCA representation is robust in the presence of noise, and remains more robust to different sampling rates than unwarped trajectories.}}
\label{fig:meta}
\vspace{-10pt}
\end{figure*}

A potential weakness in the present TSRVF framework is in the choice of the reference point $c$, which may introduce unwanted distortions if chosen incorrectly. In manifolds such as the $SE(3)$ and $SPD$, a natural candidate for $c$ is $I_4$, however for other manifolds such as the Grassmann, the reference must be chosen experimentally. In such cases, a common solution is to choose the Riemannian center of mass (RCM), since it is equally distant from all the points thereby minimizing the possible distortions. In our experiments we show that choosing an arbitrarily bad reference point can lead to poor convergence when warping multiple trajectories. 
We test the stability of the TSRVF representation to the choice of reference point by studying the convergence rate. We chose a set of $10$ similar actions from the UMD actions dataset and measured registration error over time. The registration error is measured as $\Sigma_j d(\mu(t)-\alpha_j(t))^2$, where $\mu(t)$ is the current estimate of the mean as described in algorithm \ref{algo:seq_pca}. When $c$ is chosen as the mean, the convergence occurs in about $35$ iterations as seen in \ref{fig:stability}. To generate an arbitrary reference point, we chose a point at random from the dataset and travel along an arbitrary direction from that point. The resulting point is chosen as the new reference point and the unwarped trajectories are now aligned by transporting the TSRVFs to the new $c$. In order to account for the randomness, we repeat this experiment $10$ times and take the average convergence error. The distortion is clearly visible in figure \ref{fig:stability}, where there is no sign of convergence even after $80$ iterations. 

\subsection{Effect of Noise}
In the Euclidean setting, the robustness of PCA to noisy data is well known. We examine the consequences of performing PCA on noisy trajectories for activity recognition here. There are many different stages of adding noise to a trajectory in this context - a) sensor noise which is obtained due to poor background segmentation or sensor defect that causes the resulting shape feature to be distorted, b) warping noise that is caused by a poor warping algorithm and c) TSRVF noise, which is obtained due to a poor choice of the reference point, or obtained as a consequence to parallel transport. We have studied the effect of the reference point previously, and the effect of poor warping is unlikely in realistic scenarios. We consider the noise at the sensor level which is most likely, by inducing noise in the shape feature. We perform this by perturbing each shape point on the Grassmann manifold along a random direction, $v_r \in \mathcal{T}_{\alpha(i)}(\mathcal{G})$, for a random distance, $k$ drawn from a uniform distribution: $k \in \mathcal{U}(0,1)$. We generate the random tangent and the random distance to be traversed uniformly. Therefore, the $i^{th}$ point in a trajectory is transformed as : $\hat{\alpha}(i) = \mbox{\textbf{exp}}(\alpha(i),k~v_r)$. We then perform a recognition experiment on the noisy datasets using the RFPCA, TSRVF and unwarped representations. Figure \ref{fig:noise} shows the results of the experiment on the UMD actions dataset, with $k$ on the X-axis. As expected, the RFPCA representation is least affected, while the TSRVF representation performs slightly better than the unwarped trajectories. The different levels of noise indicate how far along the random vector one traverses to obtain the new \emph{noisy} shape.

\subsection{Arbitrary length \& sampling rates}
The choice of parameter $T$ in algorithm \ref{algo:seq_pca}, directly affects the resulting dimensionality of the the trajectory before performing coding. Here we investigate its effect on coding and recognition. We can generate different trajectory lengths by considering two factors a)frame-rate, where $\widehat{\alpha}(t) = \alpha(\mbox{\textbf{m}}t)$ where the factor is governed by $\mbox{\textbf{m}}$, and b) arbitrary end point, where $\widehat{\alpha}(t) = \alpha(1:T')$, such that $T'<T$. The TSRVF is invariant to frame rate or sampling rate, therefore for a wide range of sampling rates, the recognition accuracy remains unchanged. To observe this, we perform a recognition experiment on the Florence3D skeleton actions dataset. The results for both factors are shown in figure \ref{fig:sampling}, and it is seen that in both cases the TSRVF warped actions are recognized better than the unwarped actions with an average of ~$5 \%$ better accuracy. 

\noindent {\bf Canonical length:} Using the coding framework proposed in this paper, it is conceivable that there is a close relationship between the \emph{true} length of a trajectory and its intrinsic dimensionality. For example - a more complex trajectory contains more information which naturally requires a higher dimensional RFPCA space to truly capture its variability. However, determining the explicit relationship between the RF-PCA dimension and the canonical length of a trajectory is out of the scope of this work. 

\section{Conclusion \& Future Work}
\label{sec:conc}
In this paper we introduced techniques to explore and analyze sequential data on Riemannian manifolds, applied to human activities, visual speech recognition, and stroke rehabilitation. We employ the TSRVF space \cite{Jingyong2014}, which provides an elastic metric between two trajectories on a manifold, to learn the latent variable space of actions, which is a generalization of manifold learning to Riemannian trajectories. We demonstrate these ideas on the curved product space $SE(3) \times \dots \times SE(3)$ for skeletal actions, the Grassmann manifold, and the SPD matrix manifold. We propose a framework that allows for the parameterization of Riemannian trajectories using popular coding methods -- RF-PCA which generalizes functional PCA to manifolds and PGA to sequences, sparsity inducing coding RF-KSVD and discriminative RF-LCKSVD. The learned codes not only provide a compact and robust representation that outperforms many state of the art methods, but also the visualization of actions due to its ability to reconstruct original non-linear features. We also show applications for intuitive visualization of abstract properties such as quality of movement, which has a proven benefit in rehabilitation. The proposed representation also opens up several opportunities to understand various properties of Riemannian trajectories, including their canonical lengths, their intrinsic dimensionality, ideal sampling rates, and other inverse problems which are sure to benefit several problems involving the analysis of temporal data.

\vspace{-10pt}
\bibliographystyle{ieee}
\bibliography{ref1,ref2}
\vspace{-40pt}

\end{document}